\documentclass[11pt]{article}

\usepackage{xcolor}         
\usepackage{graphicx}
\usepackage{wrapfig}
\usepackage{tabu}
\usepackage{url}
\usepackage{amsmath}
\usepackage{amssymb}
\usepackage{amsthm}
\usepackage{algorithm}
\usepackage{subfigure}
\usepackage{multirow}
\usepackage{tcolorbox}
\usepackage{natbib} 

\usepackage{hyperref}
\usepackage{algpseudocode}
\algnewcommand\algorithmicinput{\textbf{Input:}}
\algnewcommand\algorithmicoutput{\textbf{Output:}}
\algnewcommand\Input{\item[\algorithmicinput]}%
\algnewcommand\Output{\item[\algorithmicoutput]}%
\algnewcommand{\algorithmicpreprocess}{\textbf{Pre-process}}
\algnewcommand{\algorithmicoffline}{\textit{Offline}}
\algdef{SE}{preproc}{endpreproc}{\algorithmicpreprocess (\algorithmicoffline)}{\algorithmicend \ \algorithmicpreprocess}%
\algnewcommand{\algorithmicinference}{\textbf{Inference}}
\algdef{SE}{infer}{endinfer}{\algorithmicinference}{\algorithmicend \ \algorithmicinference}%
\newtheorem{assump}{Assumption}

\newtheorem{observation}{Observation}

\title{Out-of-Distribution Detection in Time-Series Domain: A Novel Seasonal Ratio Scoring Approach}
\author{Taha Belkhouja$^*$ \hspace{6cm} \hfill taha.belkhouja@wsu.edu  \\
		Yan Yan$^*$ \hspace{6cm} \hfill yan.yan1@wsu.edu@wsu.edu  \\
		Janardhan Rao Doppa$^*$ \hspace{6cm} \hfill jana.doppa@wsu.edu  \\
		$^*$School of EECS, Washington State University \hspace{7cm}\hfill }
		
\date{}
\begin{document}

\maketitle

\begin{abstract}
Safe deployment of time-series classifiers for real-world applications relies on the ability to detect the data which is not generated from the same distribution as training data. This task is referred to as out-of-distribution (OOD) detection. We consider the novel problem of OOD detection for the time-series domain. We discuss the unique challenges posed by time-series data and explain why prior methods from the image domain will perform poorly. Motivated by these challenges, this paper proposes a novel {\em Seasonal Ratio Scoring (SRS)} approach. SRS consists of three key algorithmic steps. First, each input is decomposed into class-wise semantic component and remainder. Second, this decomposition is employed to estimate the class-wise conditional likelihoods of the input and remainder using deep generative models. The seasonal ratio score is computed from these estimates. Third, a threshold interval is identified from the in-distribution data to detect OOD examples. Experiments on diverse real-world benchmarks demonstrate that the SRS method is well-suited for time-series OOD detection when compared to baseline methods. Open-source code for SRS method is provided at \href{https://github.com/tahabelkhouja/SRS}{https://github.com/tahabelkhouja/SRS}.
\end{abstract}

\section{Introduction}
Time-series data analytics using deep neural networks \cite{ismail2019deep} enable many real-world applications in health-care \citep{tripathy2018use}, finance \citep{yang2021novel}, smart grids \citep{zheng2017wide}, energy management \cite{EM1,EM2}, and intrusion detection \citep{kim2017method}. However, safe and reliable deployment of such machine learning (ML) systems require robust models \cite{TCAD,JAIR,ICCAD,TPAMI,AAAI2022}, uncertainty quantification \cite{NCP,PRCP}, and the ability to detect time-series data which do not follow the distribution of training data, aka in-distribution (ID). For example, a circumstantial event for an epilepsy patient or a sudden surge in a smart grid branch result in sensor readings which deviate from the training distribution. This task is referred to as out-of-distribution (OOD) detection. If the ML model encounters OOD inputs, it can output wrong predictions with high confidence.  Another important application of OOD detection for the time-series domain is synthetic data generation. Many time-series applications suffer from limited or imbalanced data, which motivates methods to generate synthetic data \citep{smith2020conditional}. A key challenge is to automatically assess the validity of synthetic data, which can be alleviated using accurate OOD detectors.  

There is a growing body of work on OOD detection for the image domain \citep{hendrycks2017baseline,liu2020energy,liang2017enhancing,xiao2020regret,YY2020roblocallip,cao2020benchmark} and other types of data such as genomic sequences \citep{jie2019likelihood}. These methods can be categorized into
\begin{itemize}
    \item Supervised methods which fine-tune the ML system or perform specific training to distinguish examples from ID and OOD.
    \item Unsupervised methods which employ Deep Generative Models (DGMs) on unlabeled data to perform OOD detection.
\end{itemize}

However, time-series data with its unique characteristics (e.g., sparse peaks, fast oscillations) pose unique challenges that are not encountered in the image domain: 
\begin{itemize}
    \item Spatial relations between pixels are not similar to the temporal relations across different time-steps of time-series signals.
    \item Pixel variables follow a categorical distribution of values $\{0,1,\cdots,255\}$ where as time-series variables follow a continuous distribution.
    \item The semantics of images (e.g., background, edges) do not apply to time-series data.
    \item Humans can identify OOD images for fine-tuning purposes, but this task is challenging for time-series data. Hence, prior OOD methods are not suitable for the time-series domain.
\end{itemize}

\begin{figure*}[!t]
    \centering  
    \includegraphics[width=\linewidth]{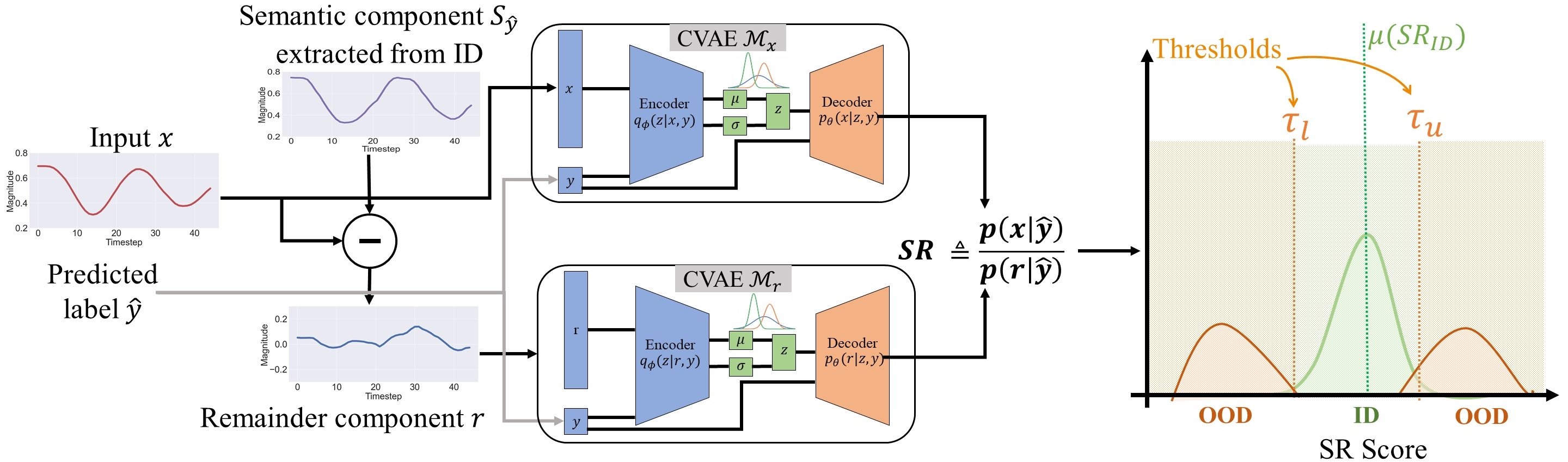}
    \caption{Overview of the seasonal ratio (SR) scoring algorithm.  
    The semantic component $S_{\hat{y}}$ for the predicted output $\hat{y}$ is obtained from the training stage via Seasonal and Trend decomposition using Loess (STL). The semantic component $S_{\hat{y}}$ is subtracted from the time-series  $x$ to obtain the remainder $r$. The trained CVAE models $\mathcal{M}_x$ and $\mathcal{M}_r$ are used to compute the SR score. If the SR score is within the threshold interval $[\tau_l, \tau_u]$ identified during training, then $x$ is classified as ID. Otherwise, it is flagged as OOD.}
    \label{fig:framework}
\end{figure*}

This paper proposes a novel OOD detection algorithm for the time-series domain referred to as Seasonal Ratio Scoring (SRS). {\em To the best of our knowledge, this is the first work on OOD detection over time-series data}. 
SRS employs the Seasonal and Trend decomposition using Loess (STL) \citep{cleveland1990stl} on time-series signals from the ID data to create a class-wise semantic pattern and a remainder component for each signal. For example, in a human activity recognition system, SRS would extract a pattern "running" that describes semantically all the recorded "running" windows. If the person trips and falls, SRS would detect that this event does not belong to the pre-defined activity classes and flag it as OOD. For this purpose, we train two separate DGMs to estimate the class-wise conditional likelihood of a given time-series signal and its STL-based remainder component. The Seasonal Ratio (SR) score for each time-series signal from ID is computed from these two estimates. A threshold interval is estimated from the statistics of all these scores over ID data. Given a new time-series input and a classifier at the testing time, the SRS approach computes the SR score for the predicted output and flags the time-series signal as OOD example if the score lies outside the threshold interval. Figure~\ref{fig:framework} illustrates the SRS algorithm. The effectiveness of SRS critically depends on the extraction of accurate class-wise semantic components. Since time-series data is prone to warping and time-shifts, we also propose a new alignment approach based on Dynamic Time Warping (DTW) \citep{muller2007dynamic} to improve the output accuracy of STL decomposition.  Our experiments on diverse real-world time-series datasets demonstrate that SRS method is well-suited for time-series OOD detection when compared to prior methods.

\vspace{1.0ex}

\noindent {\bf Contributions.} The main contribution of this paper is the development and evaluation of the Seasonal Ratio Scoring (SRS) algorithm for OOD detection in the time-series domain. Specific contributions include:
\begin{enumerate}
    \item Principled algorithm based on STL decomposition and deep generative models to compute the Seasonal Ratio (SR) score to detect OOD time-series examples.
    \item Novel time-series alignment algorithm based on dynamic time warping to improve the effectiveness of the SR score-based OOD detection.
    \item Formulation of the experimental setting for time-series OOD detection. Experimental evaluation of SRS algorithm on real-world datasets and comparison with state-of-the-art baselines.
    \item Open-source code and data for SRS method \href{https://github.com/tahabelkhouja/SRS}{https://github.com/tahabelkhouja/SRS}.
\end{enumerate}

\section{Problem Setup}
Suppose $\mathcal{D}_{in}$ is an in-distribution (ID) time-series dataset with $d$ examples $\{(x_i, y_i)\}$ sampled from the distribution $P^*$  defined on the joint space of input-output pairs $(\mathcal{X}, \mathcal{Y})$. Each $x_i \in \mathbb{R}^{n \times T}$  from $\mathcal{X}$ is a multi-variate time-series input, where $n$ is the number of channels and $T$ is the window-size of the signal. $y_i \in \mathcal{Y}=\{1,\cdots,C\}$ represents the class label for time-series signal $x_i$. 
We consider a time-series classifier $F: \mathbb{R}^{n \times T} \rightarrow \{1, \cdots, C\}$ learned using $\mathcal{D}_{in}$.  For example, in a health monitoring application using physiological sensors for patients diagnosed with cardiac arrhythmia, we use the measurements from wearable devices to predict the likelihood of cardiac failure.

OOD samples $(x,y)$ are typically generated from a distribution other than $P^*$. Specifically, we consider a sample $(x,y)$ to be OOD if the class label $y$ is different from the set of in-distribution class labels, i.e., $y \notin \mathcal{Y}$. The classifier $F_{\theta}(x)$ learned using $\mathcal{D}_{in}$ will assign one of the $C$ class labels from $\mathcal{Y}$ when encountering an OOD sample $(x,y)$. Our goal is to detect such OOD examples for safe and reliable real-world deployment of time-series classifiers.
We provide a summary of the mathematical notations used in this paper in Table \ref{tab:notation}.

\begin{table}[h]
\caption{Mathematical notations used in this paper.}
\label{tab:notation}
    \centering
\resizebox{.8\linewidth}{!}{%
    \centering
    \begin{tabular}{|c|l|}
    \hline
\textbf{Variable}  & \textbf{Definition} \\ \hline
$\mathcal{D}_{in}$ & dataset of in-distribution time-series signals\\ \hline
$P^*$ & True distribution of the time-series dataset \\ \hline
$x_i$ & Input time-series signal \\ \hline
$\mathcal{Y}$ & Set of output class labels $y \in \{1,\cdots,C\}$ \\ \hline
$F_{\theta}$ & Classifier that maps an input $x_i\in \mathbb{R}^{n \times T}$ to a class label $ y\in \mathcal{Y}$ \\ \hline
$S_y$ & Semantic pattern of a class $y$ according to STL decomposition\\ \hline
$SR$ & Seasonal Ratio score \\ \hline
$SRS$ & Seasonal Ratio Scoring framework \\ \hline
\end{tabular}
}
\end{table}
\noindent \textbf{Challenges of time-series data.} The unique characteristics of time-series data (e.g., temporal relation across time-steps, fast oscillations, continuous distribution of variables) pose unique challenges not seen in the image domain. Real-world time-series datasets are typically small (relative to image datasets) and exhibit high class-imbalance \cite{dau2019ucr}. Therefore, estimating a good approximation of in-distribution $P^*$ is hard, which results in the failure of prior OOD methods. Indeed, our experiments demonstrate that prior OOD methods are not suited for the time-series domain. As a prototypical example, Figure \ref{fig:histoverlap} shows the limitation of Likelihood Regret score \citep{xiao2020regret} to identify OOD examples: ID and OOD scores of real-world time-series examples overlap.

\section{Background and Preliminaries}

In this section, we provide the necessary background on conditional VAE and STL decomposition, to better understand the proposed Seasonal Ratio (SR) score-based OOD detection approach.

\vspace{1.0ex}

\noindent {\bf Conditional VAE.} Variational Auto-Encoders (VAEs) are a class of likelihood-based generative models with many real-world applications \citep{doersch2016tutorial}. They rely on the encoding of a raw input $x$ as a latent Gaussian variable $z$ to estimate the likelihood of $x$. The latent variable $z$ is used to compute the likelihood of the training data: $p_{\theta}(x)=\int p_{\theta}(x|z)p(z)dz$. Since the direct computation of this likelihood is impractical, the principle of evidence lower bound (ELBO) \citep{xiao2020regret} is employed. In this work, we consider the ID data $\mathcal{D}_{in}$ as $d$ input-output samples of the form $(x_i, y_i)$. We want to estimate the ID using both $x_i$ and $y_i$. Therefore, we propose to use conditional VAE (CVAE) for this purpose. CVAEs are a class of likelihood-based generative models  \citep{doersch2016tutorial}. They rely on the encoding of raw input $(x,y)$ as a latent Gaussian variable $z$ to estimate the conditional likelihood of $x$ over the class label $y$. CVAE is similar to VAE with the key difference being the use of conditional probability over both $x_i$ and $y_i$. 
The ELBO objective of CVAE is:
\begin{equation*}
    \mathcal{L}_{\text{ELBO}}\overset{\Delta}{=}  \mathbb{E}_{\phi}\big[\log p_{\theta}(x|z,y)\big]  -  D_{\text{KL}}\big[q_{\phi}(z|x,y)||p(z|y)\big]
\end{equation*}
where $q_{\phi}(z|x,y)$ is the variational approximation of the true posterior distribution $p_{\theta}(x|z,y)$. As CVAE only computes the lower bound of the log-likelihood of a given input, the exact log-likelihood is estimated using Monte-Carlo sampling as shown below:
\begin{equation}
    \mathcal{L}_M=\mathbb{E}_{z^m \sim q_{\phi}(z|x,y)}\bigg[\log \dfrac{1}{M} \sum_{m=1}^M \dfrac{p_{\theta}(x|z^m,y)p(z^m)}{q_{\phi}(z^m|x,y)}\bigg]
    \label{eq:cvae_mc}
\end{equation}

The intuitive expectation from a DGM learned using training data is to assign a high likelihood to ID samples and a low likelihood to OOD samples. However, recent research showed that DGMs tend to assign highly unreliable likelihood to OOD samples regardless of the different semantics of both ID and OOD data \citep{xiao2020regret}. Indeed, our experimental results shown in Table \ref{tab:OODrecons} demonstrate that this observation is also true for the time-series domain.

\vspace{1.0ex}

\begin{figure}[t]
\centering
    \begin{minipage}{.38\linewidth}
        \centering
        \includegraphics[width=\linewidth]{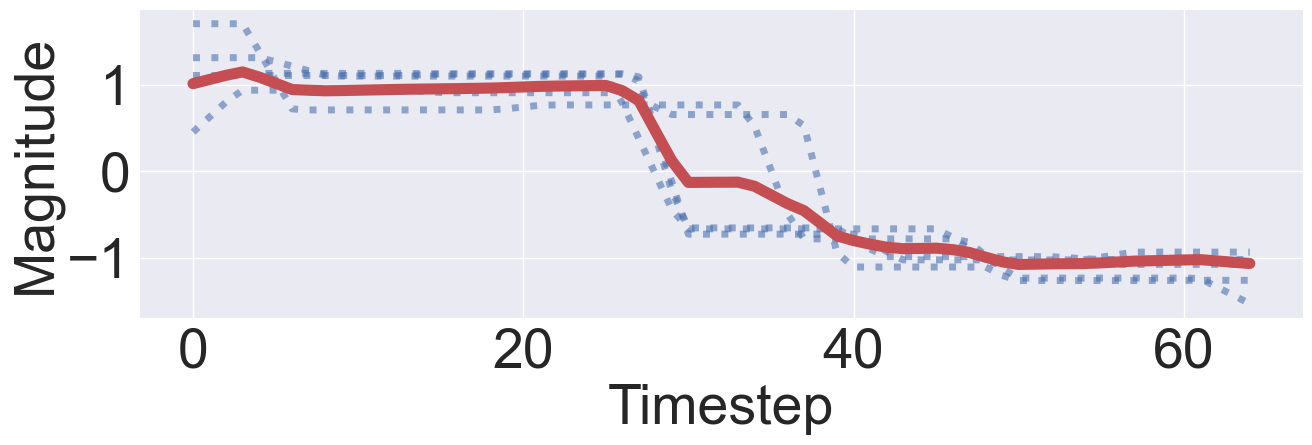}
    \end{minipage}%
    \hspace{2ex}
    \begin{minipage}{.38\linewidth}
        \centering
        \includegraphics[width=\linewidth]{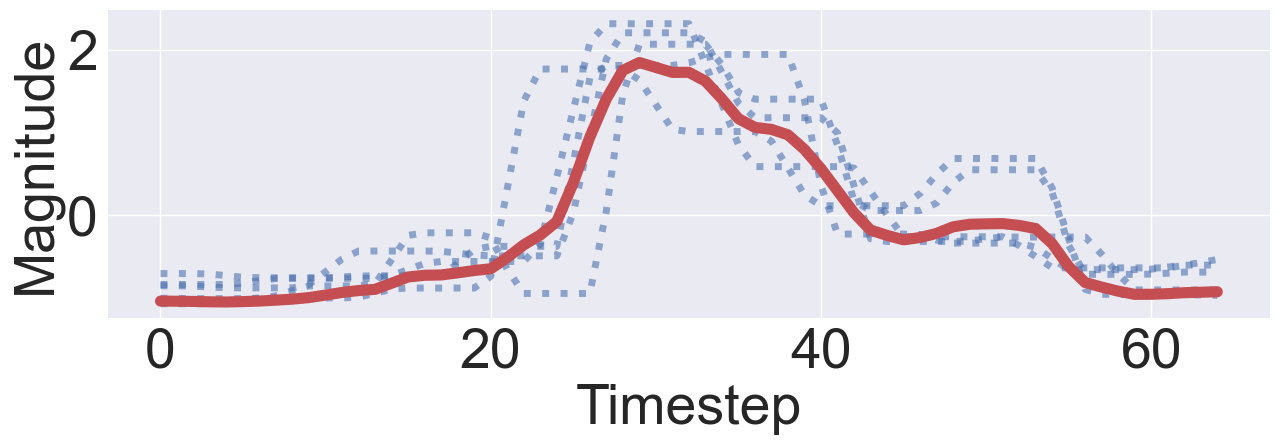}
    \end{minipage}
    \caption{Illustration of STL method for two different classes from the ERing dataset. Dotted signals are natural time-series signals $x$ and the red signal is the semantic pattern $S_y$.}
    \label{fig:stlexample}
\end{figure}

\noindent \textbf{STL decomposition.}  STL \citep{jiang2011part} is a statistical method for decomposing a given time-series signal $x$ into three different components: 1) The seasonality $x_s$ is a fixed regular pattern that recurs in the data; 2) The trend $x_t$ is the increment or the decrement of the seasonality over time; and 3) The residual $x_r$ represents random additive noise. STL employs Loess (LOcal regrESSion) smoothing in an iterative process to estimate the seasonality  component $x_s$ \citep{mckinney2011time}. The remainder is the additive residual from the input $x$ after summing both $x_s$ and $x_t$. For the proposed SRS algorithm, we assume that there is a fixed semantic pattern $S_{y}$ for every class label $y \in \mathcal{Y}$, and this pattern recurs in all examples $(x_i, y_i)$ from $\mathcal{D}_{in}$ with the same class label, i.e., $y_i$=$y$. We will elaborate more on this assumption and a reformulation of the problem that can be used when the assumption is violated in the next section. Hence, every time-series example has the following two elements: $x_i = S_{y_i} + r_i$, where $S_{y_i}$ is the pattern for the class label $y$=$y_i$ and $r_i$ is the remainder noise w.r.t $S_{y_i}$. For time-series classification tasks, ID samples are assumed to be stationary. Therefore, we propose to average the trend $x_t$ component observed during training and include it in the semantic pattern $S_{y}$. Figure \ref{fig:stlexample} illustrates the above-mentioned decomposition for two different classes from the ERing dataset.

\section{Seasonal Ratio Scoring Approach for OOD Detection}

\vspace{1.0ex}

\noindent {\bf Overview of SRS algorithm.} The training stage proceeds as follows. We employ STL decomposition to get the semantic component $S_y$ for each class label $y \in \mathcal{Y}$=$\{1,\cdots,C\}$ from the given in-distribution (ID) data $\mathcal{D}_{in}$. Details on the STL decomposition steps within the SRS framework are provided in Section \ref{sec:srs_steps}. To improve the accuracy of STL decomposition, we apply a time-series alignment method based on dynamic time warping to address scaling, warping, and time-shifts as detailed in Section \ref{sec:srs_alignment}. We train two CVAE models $\mathcal{M}_x$ and $\mathcal{M}_r$ to estimate the class-wise conditional likelihood of each time-series signal $x_i$ and its remainder component $r_i$ w.r.t the semantic component $S_{y_i}$. The seasonal ratio score for each ID example $(x_i, y_i)$ from $\mathcal{D}_{in}$ is computed as the ratio of the class-wise conditional likelihood estimates for $x_i$ and its remainder $r_i$:  $SR_i(x_i, y_i) \overset{\Delta}{=} \frac{p(x_i|y=y_i)}{p(r_i|y=y_i)}$. We compute the SR scores for all in-distribution examples from $\mathcal{D}_{in}$ to estimate the threshold interval $[\tau_l, \tau_u]$ for OOD detection. During the inference stage, given a time-series signal $x$ and a trained classifier $F(x)$, we compute the SR score of $x$ with the predicted output $\hat{y}$=$F(x) \in \mathcal{Y}$ and identify it as an OOD example if the SR score lies outside the threshold interval $[\tau_l, \tau_u]$. Figure \ref{fig:framework} provides a high-level illustration of the SRS algorithm.

\vspace{1.0ex}

Below we first provide an intuitive explanation to motivate the SR score. Next, we describe the complete details of the SRS algorithm including both training and inference stages. Finally, we motivate and describe a time-series alignment approach based on dynamic time warping to improve the effectiveness of SRS.

\subsection{Intuition for Seasonal Ratio Score}

We explain the intuition behind the proposed SRS algorithm using STL decomposition of time-series signals and CVAE models for likelihood estimation. Current research shows that DGMs alone can fail to identify OOD samples \citep{xiao2020regret}. They not only assign high likelihood to OOD samples, but they also exhibit good reconstruction quality. In fact, we show in Table \ref{tab:OODrecons} that CVAEs trained on a given ID data generally exhibit a low reconstruction error on most of the OOD samples. Furthermore, we show in Table \ref{tab:auroc} that using a trained CVAE likelihood output for OOD detection fails to perform well. These results motivate the new for a new OOD scoring method for the time-series domain.

\vspace{1.0ex}

\noindent {\bf Class-wise seasonality via STL decomposition.} The proposed SRS algorithm relies on the following assumption to analyze the time-series space for OOD detection.
\bigbreak
\fbox{%
\begin{minipage}{.96\linewidth}
\begin{assump}Each time-series example $(x_i, y_i)$ from the in-distribution data $\mathcal{D}_{in}$ consists of two components. 1) A class-wise semantic pattern $S_y$ for each class label $y \in \mathcal{Y}$ representing the meaningful semantics of the class label $y$. 2) A remainder noise $r_i$ representing an additive perturbation to the semantic portion. Hence, $\forall (x_i, y_i) \in \mathcal{D}_{in}:~ x_i = S_{y_i} + r_i$
\label{asp:main}%
\end{assump}%
\end{minipage}
}%
\bigbreak
\noindent We propose to employ STL decomposition to estimate semantic pattern $S_y$ (as illustrated in Figure \ref{fig:stlexample}) and deduce the remainder noise that can be due to several factors including errors in sensor measurements and noise in communication channels. These two components are analogous to the foreground and the background of an image, where the foreground is the interesting segment of the input that describes it, and the background may not necessarily be related to the foreground. In spite of this analogy,
prior methods for the image domain are not suitable for time-series as explained in the related work (Section \ref{sec:related_work}). In this decomposition, we cannot assume that $S_y$ and $r$ are independent for a given time-series example $(x, y)$, as $S_y$ is class-dependant and $r$ is the remainder of the input $x$ using $S_y$. Hence, we present their conditional likelihoods in the following observation.
\begin{observation}
Let $x \in \mathbb{R}^{n \times T}$ is a time-series signal and $y_i \in \mathcal{Y}$=$\{1,\cdots,C\}$ be the corresponding class label. As $x$ = $S_{y_i} + r$, we have: 
\begin{equation}
p(x|y_i) = p(r|y_i)p(S_{y_i}|y_i)
\end{equation}
\label{th:likelihood}
\end{observation}

\textbf{Proof of Observation 1}
As $X = S_{y_i} + r$, it is intuitive to think that $p(X|y=y_i) = p(S_{y_i})\times p(r)$. However, we cannot assume that $S_{y_i}$ and $r$ are independent, as $S_{y_i}$ is class-dependant and $r$ is the remainder of the input $X$ given $S_{y_i}$. 

\vspace{1.0ex}

\noindent Therefore, we make use of the conditional probabilities of the components. The likelihood $p(X)$ can be decomposed as follows:
\begin{equation*}
\begin{split}
p(X|y=y_i) &= p(S_{y_i},r|y=y_i)\\
     &= \frac{p(S_{y_i},r,y=y_i)}{p(y=y_i)}\\
     &= \frac{p(r,y=y_i)p(S_{y_i}|r,y=y_i)}{p(y=y_i)}
\end{split}
\end{equation*}
For the conditional probability $p(S_{y_i}|r,y=y_i)$, as only the pattern $S_{y_i}$ depends on the class label, and that we have defined $r$ as a non-meaningful noise to the input, we can assume that \textbf{$S_{y_i}$ and $r$ are conditionally independent given the class $y_i$}. Therefore, we have the following.
\begin{equation*}
\begin{split}
p(X|y=y_i) &= \frac{p(r,y=y_i)p(S_{y_i}|r,y=y_i)}{p(y=y_i)}\\
     &= \frac{p(r,y=y_i)p(S_{y_i}|y=y_i)}{p(y=y_i)}\\
     &= \frac{p(r|y=y_i)p(y=y_i)p(S_{y_i}|y=y_i)}{p(y=y_i)}\\
     &= p(r|y=y_i)p(S_{y_i}|y=y_i)\\
\end{split}
\end{equation*}

\vspace{1.0ex}
\noindent \textbf{Discussion on Observation 1.} $S_y$ is a fixed class-wise semantic pattern that characterizes a class $y \in \mathcal{Y}$. By definition, $S_y$ is a deterministic pattern extracted using STL decomposition during training and is not a random variable. At the inference time, we do not estimate $S_y$ of each test input $x$, but we use $S_y$ computed during the training stage to estimate the remainder component $r$. Hence, $P(S_y|y)$ is defined as a deterministic variable and not as a density that SRS is aiming to estimate.

\noindent {\bf OOD detection using CVAEs.} Observation \ref{th:likelihood} shows the relationship between the conditional likelihood of the input $x$ and its remainder $r$. We propose to employ CVAEs to estimate both likelihoods since they are conditional likelihoods. Recall that OOD examples come from an unknown distribution which is different from the in-distribution $P^*$ and do not belong to any pre-defined class label from $\mathcal{Y}$. Therefore, we propose to use the following observation for OOD detection in the time-series domain.
\begin{observation}
Let $x \in \mathbb{R}^{n \times T}$ is a time-series signal and $y \in \mathcal{Y}$=$\{1,\cdots,C\}$ be the corresponding class label. As $x$ = $S_{y} + r$, $x$ is an OOD example if $p(x|y) \neq p(r|y)$ and an in-distribution example if  $p(x|y)$ = $p(r|y)$. 
\label{th:lemmaOOD}
\end{observation}

Observation \ref{th:lemmaOOD} shows how we can exploit the relationship between the estimated conditional likelihood of the time-series signal $x$ and its remainder $r$ to predict whether $x$ is an OOD example or not. This observation relies on the assumption that $p(S_y|y)=1$ for in-distribution data. 
For ID data, the semantic pattern $S_y$ is a class-dependant signal that defines the class label $y$.  Since the semantic component is guaranteed to be $S_y$ for any time-series example with class label $y$, we have $p(S_y|y)$=1. On the other hand, OOD examples do not belong to any class label from $\mathcal{Y}$, i.e., $p(S_y|y) \neq 1$ for any $y \in \mathcal{Y}$. To estimate $p(x|y \in \mathcal{Y})$ and $p(r|y \in {\mathcal Y})$ in observation \ref{th:lemmaOOD}, we train two separate CVAE models using the in-distribution data $\mathcal{D}_{in}$. While estimating two separate distributions can cause instability, we note that 
\begin{enumerate}
    \item During hyper-parameter tuning and the definition of the ID score range $[\tau_l, \tau_u]$, any outlier that may cause estimation instability will be omitted.
    \item In case of drastic estimation instability, both CVAEs can be tuned during training time to overcome the problem.
    \item If this instability is seen during inference time, then the SRS algorithm automatically indicates that the test example is an OOD example.
\end{enumerate}

\vspace{1.0ex}

\noindent {\bf Discussion on Assumption 1.} The paper acknowledges that this assumption may fail to hold in some real-world scenarios. However, surprisingly, our experimental results shown in Table \ref{tab:patternerr} strongly corroborate this key assumption: the distance between each time-series signal $x_i$ and its semantic pattern $S_{y_i}$ is very small. The strong OOD performance of SRS algorithm in our diverse experiments demonstrates the effectiveness of a simple approach based on this assumption.  

Suppose the assumption does not hold and some class label $y$ can possess $K > 1$ different semantics $\{S^k_y\}_{k\le K}$. If we take a human activity recognition example, it is safe to think that a certain activity (e.g., running or walking) will have $K>1$ different patterns (e.g., athletic runners vs. young runners). Therefore, the decomposition in Assumption \ref{asp:main} for a given time-series example $(x_i, y_i)$ will result into a semantic pattern describing the patterns of the different sub-categories (e.g., a pattern that describes both athletic runners and young runners). By using Lowess smoothing, the STL season extracted over a multiple-pattern class is a pattern $S_y$ that is a linear combination of $\{S^k_y\}_{k\le K}$ (for our example, it describes the combination of both athletic runs and young runs). While for an in-distribution example, $p(S_y|y)$=1 of Observation 2 will not hold, $p(S_y|y)$ is likely to be well-defined from $p(S^k_y|y)$ as $\{S^k_y\}_{k\le K}$ are fixed and natural for the class label $y$. Hence, we can still rely on the CVAEs to estimate this distribution and to perform successful OOD detection. Alternatively, we can use a simple reformulation of the problem by clustering time-series signals of a class label $y$ (for which  the assumption is not satisfied) to identify sub-classes and apply the SRS algorithm on transformed data. Since we found the assumption to be true in all our experimental scenarios (see Table \ref{tab:patternerr}), we didn't find the need to apply this reformulation. 

\subsection{OOD Detection Approach}
\label{sec:srs_steps}
One key advantage of SRS method is that it can be directly executed at inference stage and does not require additional training similar to prior VAE-based methods such as Likelihood Regret scoring. 

\vspace{1.0ex}

\noindent {\bf Training stage.} Our overall training procedure for time-series OOD detection is as follows:
\begin{enumerate}
\item Train a CVAE $\mathcal{M}_x$ using in-distribution data $\mathcal{D}_{in}$ to estimate the conditional likelihood $p(x|y \in \mathcal{Y})$ of time-series signal $x$.
\item Execute STL decomposition as follows:
\begin{enumerate}
    \item From the training data $\{(x_i, y_i)\}$, we create a group $\mathcal{D}_y=\{x_i| y_i=y\}$.
    \item We concatenate all the examples $x_i\in\mathcal{D}_y$ in a single stream of data according to the $T$ dimension. If $\mathcal{D}_y$ has $k$ examples, the output is a single stream $X_{stream}\in\mathbb{R}^{n \times (k\cdot T)}$
    \item We apply STL decomposition on the stream $X_{stream}$ by defining the pattern dimensions $S_y \in \mathbb{R}^{n \times T}$.
    \item We store the semantic component $S_y$ to be used later in estimating the remainder component for any given training example $(x,y)$: $r$ = $x - S_y$.
\end{enumerate}
\item Create the remainder for each training example $(x_i, y_i) \in \mathcal{D}_{in}$ using the patterns $S_y$ for each class label $y$ $:~ r_i = x_i - S_{y_i}$. We train another CVAE $\mathcal{M}_r$ using all these remainders to estimate the conditional likelihood $p(r|y \in {\mathcal Y})$.
\item  Compute seasonal ratio score for each $(x_i, y_i) \in D_{in}$ using the trained CVAEs $\mathcal{M}_x$ and $\mathcal{M}_r$.
\begin{equation}
    SR_i(x_i, y_i) \overset{\Delta}{=} \dfrac{p(x_i|y=y_i)}{p(r_i|y=y_i)}
\end{equation}
\item Compute the mean $\mu_{SR}$ and variance $\sigma_{SR}$ over SR scores of all in-distribution examples seen during training. Set the OOD detection threshold interval as $[\tau_l, \tau_u]$ such that $\tau_l$ = $\mu_{SR} - \lambda \times \sigma_{SR}$ and $\tau_u$ = $\mu_{SR} + \lambda \times \sigma_{SR}$, where $\lambda$ is a hyper-parameter.
\item Tune the hyper-parameter $\lambda$ on the validation data to maximize OOD detection accuracy.

\end{enumerate}

\vspace{1.0ex}

\begin{figure}[t]
\centering
    \begin{minipage}{.38\linewidth}
        \centering
        \includegraphics[width=\linewidth]{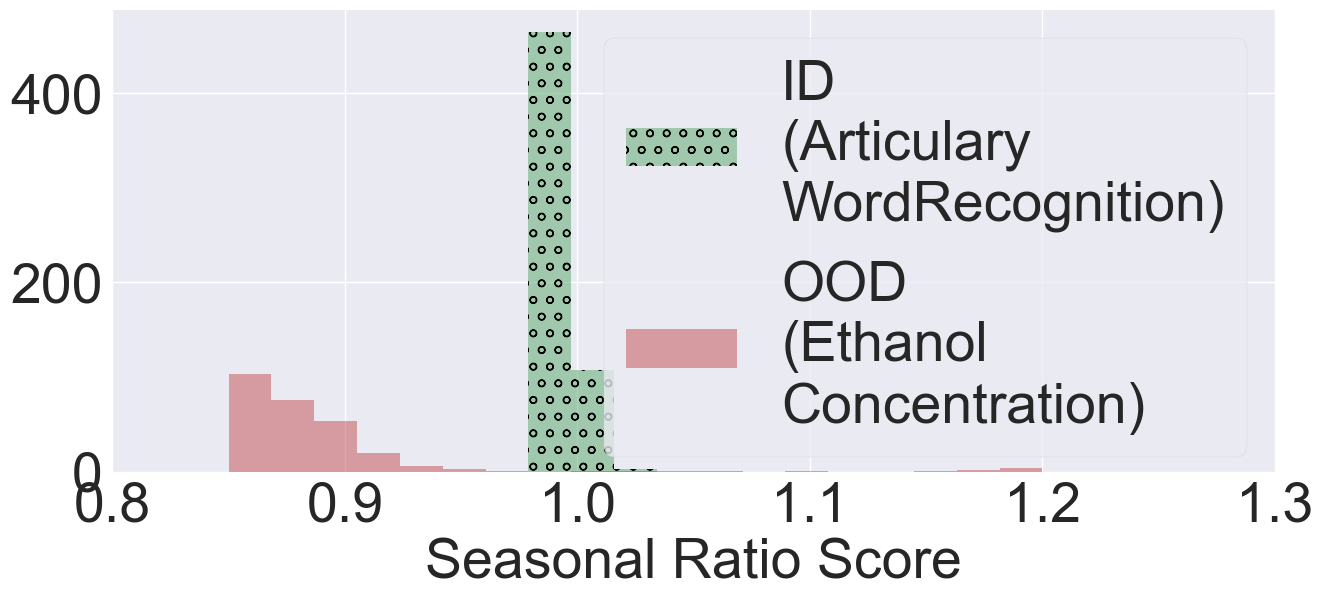}
    \end{minipage}%
    \begin{minipage}{.38\linewidth}
        \centering
        \includegraphics[width=\linewidth]{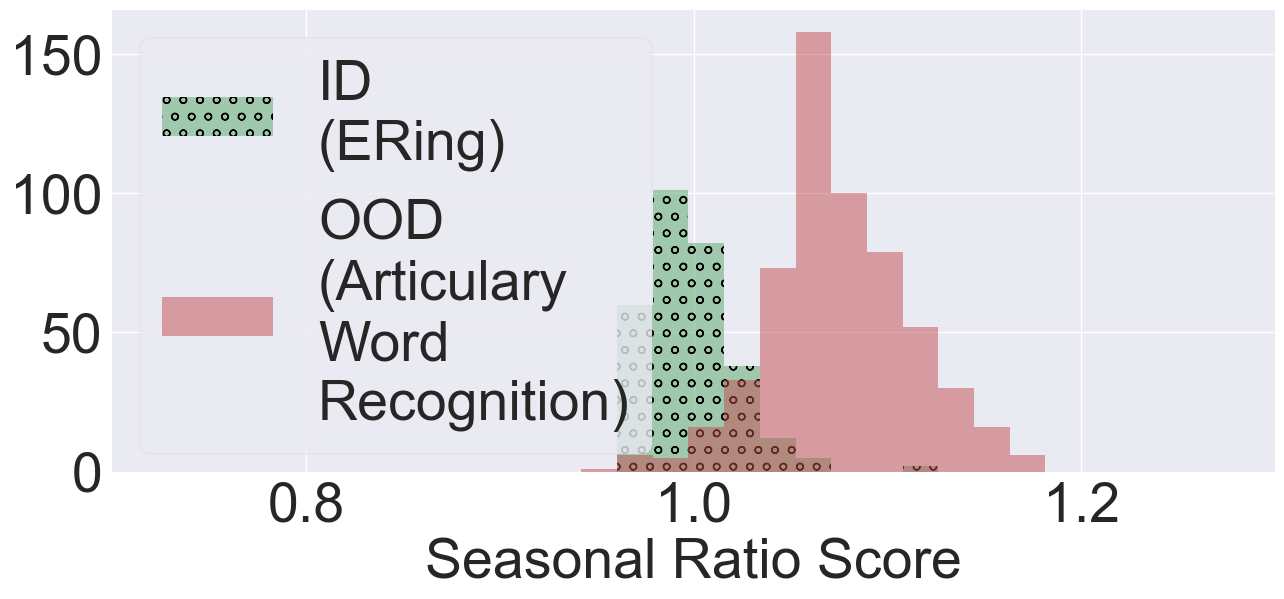}
    \end{minipage}
    \caption{Histogram showing the ID and OOD scores along the seasonal ratio score axis. The seasonal ratio scores for OOD examples can be either greater or less than the seasonal ratio scores for ID examples.}
    \label{fig:taus}
\end{figure}

The choice of $[\tau_l, \tau_u]$ for OOD detection is motivated by the fractional nature of the seasonal ratio scores. SRS algorithm assumes that in-distribution examples satisfy $p(x|y)= p(r|y)$. Hence, we characterize in-distribution examples with an SR score close to 1, whether from left ($\tau_l$) or right ($\tau_u$) side. To identify in-distribution examples, we rely on SR scores that are close to the mean score recorded during training, whether from left ($\tau_l$) or right ($\tau_u$) side. This design choice is based on the fact that SR score is a quotient ideally centered around the value 1. Indeed, we observe in Figure \ref{fig:taus} that the SR score for OOD examples can go on either the left or the right side of the SR scores for in-distribution examples. Ideally, the SR score for in-distribution examples is closer to $\mu_{SR}$ than SR scores for OOD examples, as illustrated in Figure \ref{fig:framework}. $\lambda$ is tuned to define the valid range of SR scores for in-distribution examples from $\mathcal{D}_{in}$. 
We note that the score can be changed easily to consider the quantiles of estimated ratios during training stage and use it to separate the region of OOD and ID score. Therefore, we can redefine $\tau_l = (0.5 - \lambda)$ as the $\tau_l$-quantile for the lower-limit of ID score and $\tau_u = (0.5 + \lambda)$ as the $\tau_u$-quantile for the upper-limit of ID score. Given this definition, we need to tune the hyper-parameter $0<\lambda\le 0.5$ on the validation data to maximize OOD detection accuracy. Furthermore, the ID score range is not required to be symmetric. In the general case, we can define $\tau_l = (0.5 - \lambda_l)$ as the $\tau_l$-quantile and $\tau_u = (0.5 + \lambda_u)$ as the $\tau_u$-quantile, where $\lambda_l \neq \lambda_u$. We have observed in our experiments that both these settings give similar performance. Therefore, we only consider $\tau_{u,l}$ = $\mu_{SR} \pm \lambda \times \sigma_{SR}$ for simplicity for our experimental evaluation.

\vspace{1.0ex}

\noindent {\bf Inference stage.} Given a time-series signal $x$, our OOD detection approach works as follows.
\begin{enumerate}
\item  Compute the predicted class label $\hat{y}$ using the classifier $F(x)$.
\item Create the remainder component of $x$ with the predicted label $\hat{y}$: $r$ = $x - S_{\hat{y}}$.
\item Compute conditional likelihoods $p(x|\hat{y})$ and $p(r|\hat{y})$ from trained CVAE models $\mathcal{M}_x$ and $\mathcal{M}_r$.
\item Compute the seasonal ratio score using conditional likelihoods.
\begin{equation*}
    SR(x, \hat{y})=\frac{p(x|y=\hat{y})}{p(r|y=\hat{y})}
\end{equation*}
\item If the seasonal ratio score $SR(x, \hat{y})$ does not lie within the threshold interval $[\tau_l, \tau_u]$, then classify $x$ as OOD example. Otherwise, classify $x$ as in-distribution example.
\end{enumerate}

Algorithm \ref{alg:sr} shows the complete pseudo-code including the offline training stage and online inference stage for new time-series signals. For a given time-series signal $x$ at the inference stage, we employ SRS algorithm to compute the seasonal ratio (SR) score. If the score is within $[\tau_l, \tau_u]$, then the time-series signal is classified as in-distribution. Otherwise, we flag it as an OOD time-series signal.
\begin{algorithm}[!h]
\caption{Seasonal Ratio Scoring Algorithm for OOD Detection}
\label{alg:sr}
\begin{algorithmic}[1] 
\Input $\mathcal{D}_{in} = \{(x_i, y_i)\}_{i=1}^{n}$, Input training/validation time-series samples; $\mathcal{M}_x$, CVAE for training data with likelihood $\mathcal{L}_{\mathcal{M}_x}$; $\mathcal{M}_r$, CVAE for data remainders with likelihood $\mathcal{L}_{\mathcal{M}_r}$; $F_{\theta}$, Time-series classifier; $X_{test}$, testing time-series signal;
\Output OOD Boolean decision.
\preproc
\State Cluster the time-series data in $\mathcal{D}_{in}$, one cluster for each class label ${y \in \mathcal{Y}=\{1,\cdots,C\}}$
\State Perform STL decomposition to compute class-wise semantics $\{S_y\}$
\State Compute remainder component $r_i$ for each time-series example $(x_i, y_i) \in \mathcal{D}_{in}$
\State Train CVAEs $\mathcal{M}_r$ and $\mathcal{M}_x$ using the decomposed data
\State Estimate the threshold interval $[\tau_l, \tau_u]$ from sample SR scores of in-distribution data 
\endpreproc
\State Compute the predicted class label: $\hat{y}$=$F_{\theta}(X_{test})$
\State Compute the remainder component: $r_{test}$=$X_{test}-S_{y=\hat{y}}$
\State Estimate the input log-likelihood $l_x=\mathcal{L}_{\mathcal{M}_x}(X_{test}, \hat{y})$
\State Estimate remainder log-likelihood $l_r=\mathcal{L}_{\mathcal{M}_r}(r_{test}, \hat{y})$
\State Compute the seasonal ratio score: $SR(X_{test})$ = $l_x/l_r$
\If{$SR(X_{test}) \in [\tau_l, \tau_u]$}
\State \Return FALSE
\Else
\State \Return TRUE
\EndIf
\end{algorithmic}
\end{algorithm}

\subsection{Alignment method for improving the accuracy of SRS algorithm}
\label{sec:srs_alignment}
In this section, we first motivate the need for pre-processing raw time-series signals to improve the accuracy of SRS algorithm. Subsequently, we describe a novel time-series alignment method based on dynamic time warping to achieve this goal. 

\vspace{1.0ex}

\begin{figure}[!h]
\centering
    \begin{minipage}{.4\linewidth}
        \centering
        \includegraphics[width=\linewidth]{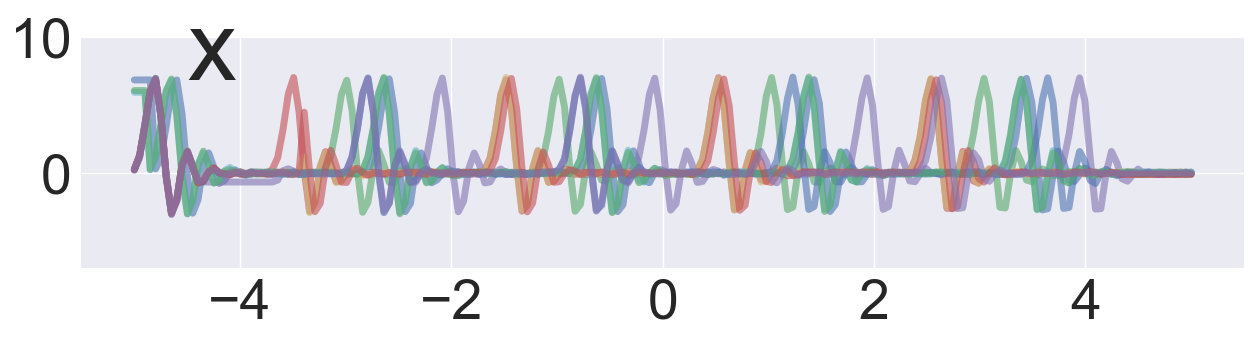}
    \end{minipage}%
    \begin{minipage}{.4\linewidth}
        \centering
        \includegraphics[width=\linewidth]{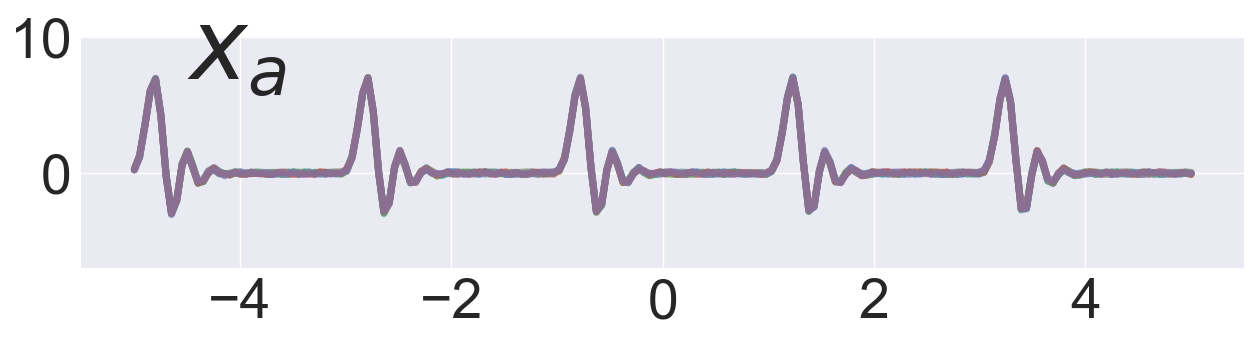}
    \end{minipage}
    \begin{minipage}{.4\linewidth}
        \centering
        \includegraphics[width=\linewidth]{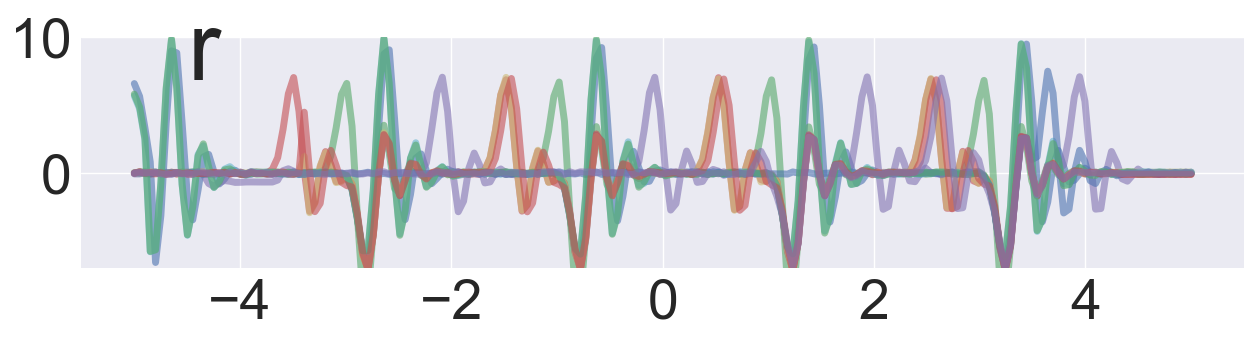}
    \end{minipage}%
    \begin{minipage}{.4\linewidth}
        \centering
        \includegraphics[width=\linewidth]{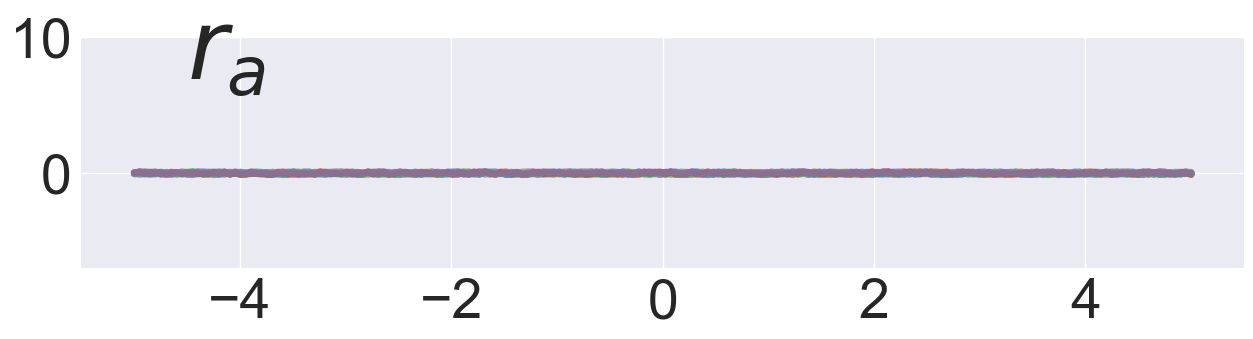}
    \end{minipage}
    \begin{minipage}{.4\linewidth}
        \centering
        Before alignment
    \end{minipage}%
    \hfill
    \begin{minipage}{.4\linewidth}
        \centering
        After alignment
    \end{minipage}
    \caption{Illustration of the challenges in time-series data for STL decomposition: semantic component and remainder. (Left column) Set of natural time-series signals with an ECG wave as semantic component $S_y$ and the corresponding remainders w.r.t $S_y$. (Right column) Time-series signals and remainders from STL decomposition after applying the alignment procedure.}
    \label{fig:patternalign}
\end{figure}

\noindent \textbf{Motivation.} The effectiveness of the SRS algorithm depends critically on the accuracy of the STL decomposition. STL method employs fixed-length window over the serialized data to estimate the recurring pattern. This is a challenge for real-world time-series signals as they are prone to scaling, warping, and time-shifts. We illustrate in Figure \ref{fig:patternalign} the challenge of scaling, warping, and time-shift occurrences in time-series data. The top-left figure depicts a set of time-series signals with a clear ECG pattern. Due to their misalignment, if we subtract one fixed ECG pattern from every time-series signal, the remainder will be inaccurate. The figures in the left and right columns show the difference in the remainder components between the {\em natural} data (Left) and the {\em aligned} version of the time-series data (Right). We can clearly observe that the remainder components from the aligned data are more accurate. If input time-series data is not aligned, it can significantly affect the estimation of $p(r_i|y=y_i)$ and the effectiveness of SRS for OOD detection. Hence, we propose a novel alignment method using the class-wise semantic for the in-distribution data $\mathcal{D}_{in}$ during both training and inference stages.

\vspace{1.0ex}

\begin{figure}[!h]
\centering
    \begin{minipage}{.29\linewidth}
        \centering
        \includegraphics[width=\linewidth]{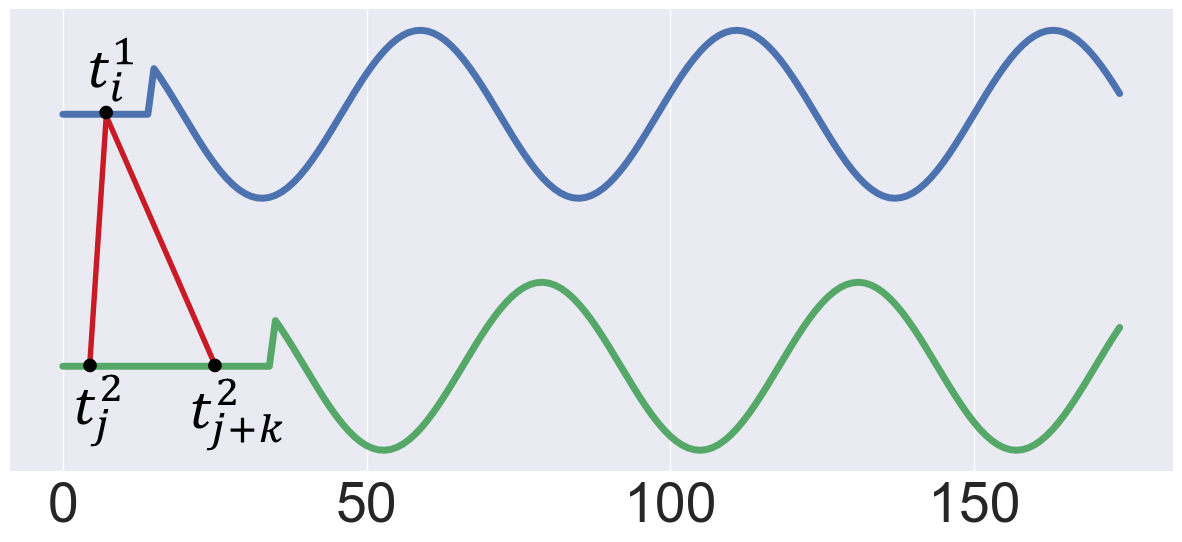}
    \end{minipage}%
    \begin{minipage}{.29\linewidth}
        \centering
        \includegraphics[width=\linewidth]{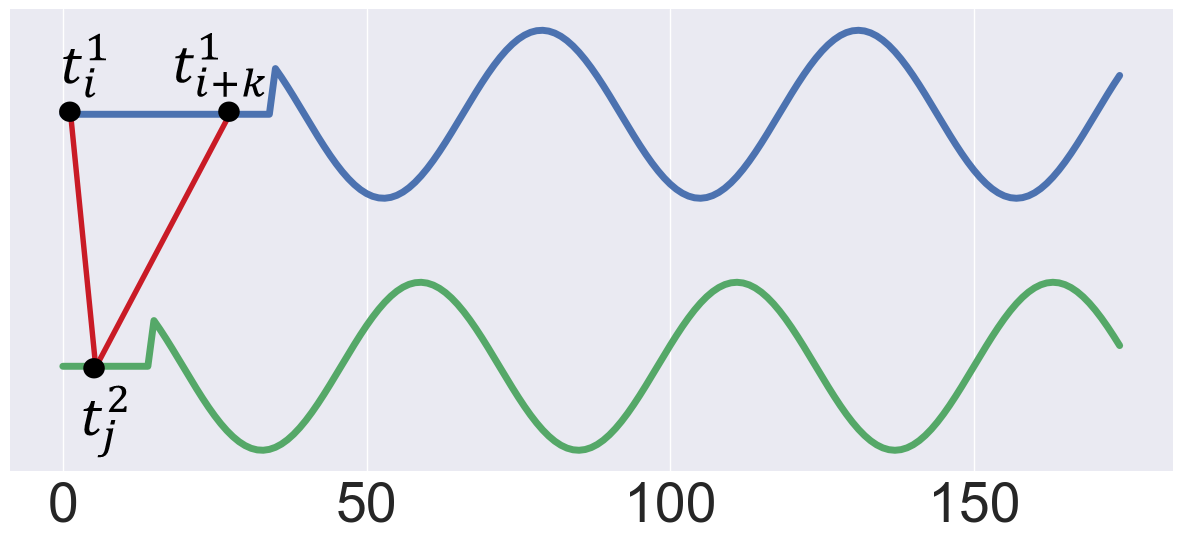}
    \end{minipage}%
    \begin{minipage}{.29\linewidth}
        \centering
        \includegraphics[width=\linewidth]{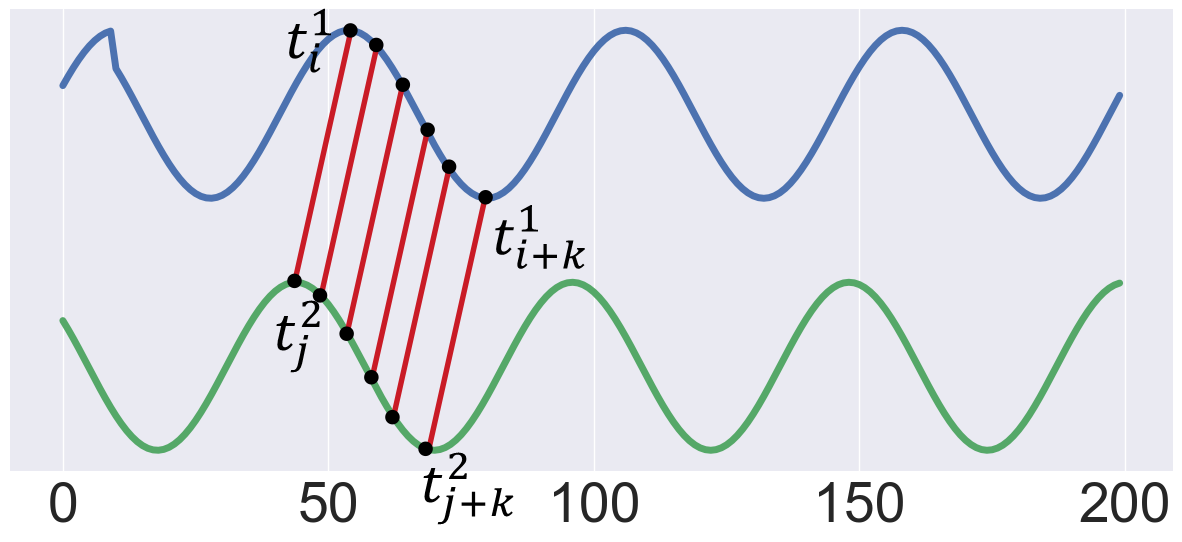}
    \end{minipage}
    \begin{minipage}{.29\linewidth}
        \centering
        (a) One-to-Many
    \end{minipage}%
    \begin{minipage}{.29\linewidth}
        \centering
        (b) Many-to-One
    \end{minipage}%
    \begin{minipage}{.29\linewidth}
        \centering
        (c) One-to-One
    \end{minipage}
    \caption{Illustration of the use of appropriate transformation to adjust the alignment between two time-series signals $X^1$ (blue signal) and $X^2$ (green signal).}
    \label{fig:talign}
\end{figure}

\noindent \textbf{Time-series alignment algorithm.} The overall goal of our approach is to produce a class-wise aligned time-series signals using the ID data $\mathcal{D}_{in}$ so that STL algorithm will produce accurate semantic components $S_{y}$ for each $y \in \mathcal{Y}$. We propose to employ dynamic time warping (DTW) \citep{muller2007dynamic} based optimal alignment to achieve this goal. The optimal DTW alignment describes the warping between two time-series signals to make them aligned in time. It overcomes warping and time-shifts issue by developing a one-to-many match over time steps. There are two key steps in our alignment algorithm. First, we compute the semantic components $S_{y}$ for each $y \in \mathcal{Y}$ from $\mathcal{D}_{in}$ using STL decomposition. For each in-distribution example $(x_i, y_i) \in \mathcal{D}_{in}$, we compute the optimal DTW alignment between $S_{y_i}$ and $x_i$. Second, we use an appropriate time-series transformation for each in-distribution example $(x_i, y_i)$ to improve the DTW alignment from the first step. Specifically, we use the time-steps of the longest one-to-many or many-to-one or sequential one-to-one sequence match to select the Expand, Reduce, and Translate transformation as illustrated in Figure \ref{fig:talign}. We define these three time-series transformations below.

\vspace{1.0ex}

\noindent  Let $X^1=(t^1_1, t^1_2, \cdots, t^1_T)$ and $X^2=(t^2_1, t^2_2, \cdots, t^2_T)$ be two time-series signals of length $T$.
\begin{itemize}
    \item  {\bf Expand$(X^1, X^2)$}: We employ this transformation for a one-to-many time-step matching ($t^1_i$ is matched with $[t^2_j, \cdots, t^2_{j+k}]$ as shown in Figure \ref{fig:talign}(a)). It duplicates the $t^1_i$ time-step for $k$ times.
    \item {\bf Reduce$(X^1, X^2)$}: We employ this transformation in the case of a many-to-one time-step matching ($[t^1_i, \cdots, t^1_{i+k}]$ is matched with $t^2_j$ as shown in Figure \ref{fig:talign}(b)). It replaces the time-steps $[t^1_i, \cdots, t^1_{i+k}]$ by a single averaged value.
    \item {\bf Translate$(X^1, X^2)$}: We employ this transformation in the case of a sequential one-to-one time-step matching ($[t^1_i, \cdots, t^1_{i+k}]$ is matched one-to-one with $[t^2_j, \cdots, t^2_{j+k}]$ as shown in Figure \ref{fig:talign}(c)). It translates $X_1$ to ensure that $t^1_i=t^2_j$.
\end{itemize}

\begin{figure}[!h]
\centering
\begin{minipage}{.8\linewidth}
    
    \begin{minipage}{.49\linewidth}
        \centering
        \includegraphics[width=\linewidth]{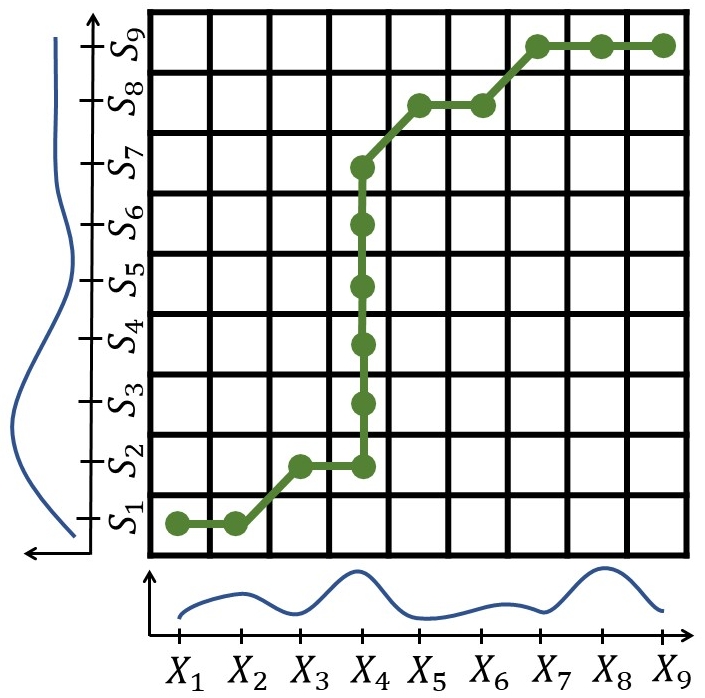}
    \end{minipage}%
    \hfill
    \begin{minipage}{.49\linewidth}
        \centering
        \includegraphics[width=\linewidth]{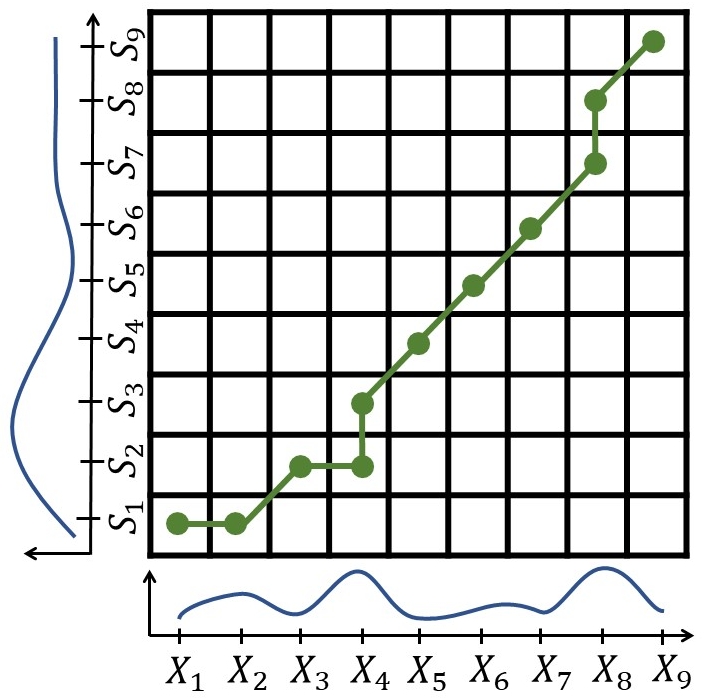}
    \end{minipage}
\end{minipage}
    \caption{Illustration of two transformation choices for a time-series $x$ aligned with a pattern $S$. (Left) One-to-many as the longest match, calling to use $Expand$ transformation. (Right) sequential one-to-one as the longest match, calling to use the $Translate$.}
    \label{fig:dtwalign}
\end{figure}

We illustrate in Figure \ref{fig:dtwalign} two examples of transformation choices for time-series signal  $x$ when aligned with a pattern $S$. The alignment on the left exhibits that the longest consecutive matching sequence is a one-to-many ($x_4$ is matched with $[S_2, \cdots, S_7]$) while the alignment on the right exhibits that the longest consecutive matching sequence is a sequential one-to-one ($[x_4, \cdots,x_8]$ is matched with $[S_3, \cdots, S_7]$).
\section{Related Work}
\label{sec:related_work}
\vspace{1.0ex}


\noindent \textbf{OOD detection via pre-trained models.} Employing pre-trained deep neural networks (DNNs) to detect OOD examples was justified by the observation that DNNs with ReLU activation can produce arbitrarily high softmax confidence for OOD examples \citep{hendrycks2017baseline}. Maximum probability over class labels has been used \citep{hendrycks2017baseline} to improve the OOD detection accuracy. Building on the success of this method, temperature scaling and controlled perturbations were used \citep{liang2017enhancing} to further increase the performance. The Mahalanobis-based scoring method \citep{lee2018unified} is used to identify OOD examples with class-conditional Gaussian distributions. Gram matrices \citep{sastry20gram} were used  to detect OOD examples based on the features learned from the training data. The effectiveness of these prior methods depends critically on the availability of a highly-accurate DNN for the classification task. However, this requirement is challenging for the time-series domain as real-world datasets are typically small and exhibit high class imbalance resulting in inaccurate DNNs \cite{wen2020time,Huang_2016_CVPR}.

\vspace{1.0ex}

\noindent \textbf{OOD detection via synthetic data.} During the training phase, it is impossible to anticipate the OOD examples that would be encountered during the deployment of DNNs \citep{hendrycks2019scaling}. Hence, unsupervised methods \citep{yu2019unsupervised} are employed or d synthetic data based on generative models is created \citep{lee2018training, lin2020using} to explicitly regularize the DNN weights over potential OOD examples. It is much more challenging to create synthetic data for the time-series domain due to the limited data and their ambiguity to be validated by human experts.

\vspace{1.0ex}

\noindent \textbf{OOD detection via deep generative models.} The overall idea of using deep generative models (DGMs) for OOD detection is as follows: 1) DGMs are trained to directly estimate the in-distribution $P^*$; and 2) The learned DGM identifies OOD samples when they are found lying in a low-density region. Prior work has used auto-regressive generative models \citep{jie2019likelihood} or GANs \citep{wang2020further} and proposed scoring metrics such as likelihood estimates to obtain good OOD detectors. DGMs are shown to be effective in evaluating the likelihood of input data and estimating the data distribution, which makes them a good candidate to identify OOD examples with high accuracy. However, as shown by \citep{nalisnick2018deep}, DGMs can assign a high likelihood to OOD examples. Likelihood ratio \citep{jie2019likelihood} and likelihood regret \citep{xiao2020regret} are proposed to improve OOD detection. While likelihood regret method can generalize to different types of data, likelihood ratio is limited to categorical data distributions with the assumption that the data contains background units (background pixels for images and background sequences for genomes). Likelihood ratio cannot be applied to the time-series domain for two reasons: 1) We need to deal with continuous distributions; and 2) We cannot assume that  time-steps (information unit) can be independently classified as background or semantic content.

\vspace{1.0ex}

\noindent \textbf{OOD detection via time-series anomaly detection.} Generic Anomaly Detection (AD) algorithms \citep{pang2021deep,ruff2021unifying,AAD} can be employed to solve OOD problems for time-series data. Anomaly detection is the task of identifying observed points or examples that deviate significantly from the rest of data. Anomaly detection relies on different approaches such as distance-based metrics or density-based approaches to quantify the dissimilarities between any example and the rest of the data. Current methods using DNNs (e.g., Generative Adversarial Networks, auto-encoders) showed higher performance in anomaly detection as they can capture more complex features in high dimensional spaces  \citep{pang2021deep,pang2019deep}. We note that there exist some AD methods that can cover the same setting as the OOD problem for time-series domain. However, both settings are still considered as two different frameworks with two different goals \cite{yang2021generalized}. By definition, AD aims to detect and flag anomalous samples that deviate from a pre-defined normality  \citep{laptev2015generic,canizo2019multi} estimated during training. Under the AD assumption of normality, such samples only originate from a covariate shift in the data distribution \citep{ruff2021unifying}. Semantically, such samples do not classify as OOD samples \citep{yang2021generalized}. For example, consider an intelligent system trained to identify the movement of a person (e.g, run, stand, walk, swim), where stumbling may occur during running. Such an event would be classified as an anomaly as the activity running is still taking place, but in an irregular manner. However, if the runner slips and falls, such activity should be flagged as OOD due to the fact that it does not belong to any of the pre-defined activity classes. In other words, OOD samples must originate from a different class distribution ($y_{\text{OOD}} \notin \mathcal{Y}$) than in-distribution examples, while anomalies typically originate from the same underlying distribution but with anomalous behavior. Open-set recognition methods can be applicable for this setting \citep{zhou2021learnings,zhou2021learning} as it has been shown that they are effective in detecting unknown categories without prior knowledge. However, OOD Detection encompasses a broader spectrum of solution space and does not require the complexity of identifying the semantic class of the anomalies. Additionally, anomalies can manifest as a single time-step, non-static window-length, but not generally a complete time-series example in itself. Such differences can be critical for users and practitioners, which necessitates the study of separate algorithms for AD and OOD. Unlike anomaly detection, OOD detection focuses on identifying test samples with non-overlapping labels with in-distribution data and can generalize to multi-class setting \citep{yang2021generalized}. The main limitations of time-series AD algorithms \cite{challu2022deep} for OOD detection tasks are
\begin{itemize}
    \item OOD samples cannot be used as labeled anomalous examples during training due to the general definition of the OOD space. For various AD methods such as nearest neighbors and distance-based, the fine-tuning of the cut-off threshold between "normal" and "anomalous" examples requires anomaly labels during training. Mainly, window-based techniques\cite{chandola2010anomaly} require both normal and anomalous sequences during training, and if there are none, anomalous examples are randomly generated. Such a requirement is not practical for OOD problem settings as the distribution is ambiguous to define and sample from.
    \item AD assumes that normal samples are homogeneous in their observations. This assumption helps the AD algorithm to detect anomalies. Such an assumption cannot hold for different classes of the in-distribution space for multi-class settings. Therefore, time-series AD algorithms are prone to fail at detecting OOD samples. Indeed, our experiments demonstrate the failure of state-of-the-art time-series AD methods.
\end{itemize}
\section{Experiments and Results}

In this section, we present experimental results comparing the proposed SRS algorithm and prior methods on diverse real-world time-series datasets.

\subsection{Experimental Setup}

\vspace{1.0ex}

\noindent \textbf{Datasets.} We employ the multivariate benchmarks from the UCR time-series repository \citep{dau2019ucr}. Due to space constraints, we present the results on representative datasets from six different pre-defined domains \textit{Motion}, \textit{ECG}, \textit{HAR}, \textit{EEG}, \textit{Audio} and \textit{Other}. The list of datasets includes Articulary Word Recognition (AWR), Stand Walk Jump (SWJ), Cricket (Ckt), Hand Movement Direction (HMD), Heartbeat (Hbt), and ERing (ERg). We employ the standard training/validation/testing splits from these benchmarks. 

\vspace{1.0ex}

\noindent \textbf{OOD experimental setting.} Prior work formalized the OOD experimental setting for different domains such as computer vision \citep{hendrycks2017baseline}. However, there is no OOD setting for the time-series domain. In what follows, we explain the challenges for the time-series domain and propose a concrete OOD experimental setting for it.

The first challenge with the time-series domain is the dimensionality of signals. Let the ID space be $\mathbb{R}^{n_i\times T_i}$ and the OOD space be $\mathbb{R}^{n_o\times T_o}$. Since we train CVAEs on the ID space, ${n_o\times T_o}$ needs to match ${n_i\times T_i}$. Hence, if $n_o>n_i$ or $T_o>T_i$, we window-clip the respective OOD dimension in order to have $n'_o=n_i$ or $T'_o=T_i$. If $n_o<n_i$ or $T_o<T_i$, we zero-pad the respective OOD dimension in order to have $n'_o=n_i$ or $T'_o=T_i$. Zero-padding is based on the assumption that the additional dimension exists but takes {\em null} values.  The second challenge is in defining OOD examples. Since the number of datasets in UCR repository is large, conducting experiments on all combinations of datasets as ID and OOD is impractical and repetitive (600 distinct configurations for the 25 different datasets considered in this paper). 

Hence, we propose two settings using the notion of domains. 
\begin{itemize}
    \item  \textbf{In-domain OOD}: Both ID and OOD datasets belong to the same domain. This setting helps in understanding the behavior of OOD detectors when real-world OOD examples come from the same application domain. For example, a detector of \textit{Epileptic} time-series signals should consider signals resulting from sports activity (\textit{Cricket}) as OOD. 
    \item \textbf{Cross-domain OOD:} Both ID and OOD datasets come from two different domains. This configuration is more intuitive for OOD detectors, where time-series signals from different application domains should not confuse the ML model (e.g., \textit{Motion} and \textit{HAR} data).
\end{itemize}

Our intuition is that the in-domain OOD setting is more likely to occur during real-world deployment. Hence, we propose to do separate experiments by treating every dataset from the same domain as OOD. For the cross-domain OOD setting, we believe that a single representative dataset from the domain can be used as OOD. In this work, we focus on real-world OOD detection for the time-series domain. Since random noise does not inherit the characteristics of time-series data, methods from the computer vision literature have a good potential in detecting random noise.

For improved readability and ease of understanding, we provide Table \ref{tab:domainlabel} and Table \ref{tab:dslabel} to explain the domain labels and dataset labels used in the experimental section of our paper along with the corresponding UCR domain name and dataset name.
\begin{itemize}
\item Table \ref{tab:domainlabel} shows the label used to represent a given domain for \textbf{Cross-Domain OOD} setting.

\begin{table}[!h]
\centering
\caption{List of domain labels used in the experimental section and the corresponding UCR domain name.}
\label{tab:domainlabel}
\begin{tabular}{|c|c|}
\hline
    \textbf{Domain label} & \textbf{Domain name} \\ \hline
    D1 & Motion \\ \hline
    D2 & ECG \\ \hline
    D3 & HAR \\ \hline
    D4 & EEG \\ \hline
    D5 & Audio \\ \hline
    D6 & Other \\ \hline
\end{tabular}
\end{table}

\item Table \ref{tab:dslabel} shows the label used to represent the dataset used as an OOD source against a given ID dataset for the \textbf{In-Domain OOD} setting.
For example, while reading Table \ref{tab:auroc} in the main paper, when AWR is the ID distribution, according to Table \ref{tab:dslabel}, DS1 represents CharacterT. dataset. On the other hand, if HMD is the ID distribution, DS1 represents FingerM. dataset. 
\end{itemize}

\begin{table*}[t]
\centering
\setlength\extrarowheight{2.5pt}
\caption{Reference table for the In-Domain dataset labels used in the experimental section and the corresponding UCR dataset name. The second column shows the average CVAE normalized reconstruction Mean Absolute Error (MAE) with a negligible variance $\le$ 0.001 on the in-distribution data.}
\label{tab:dslabel}
\resizebox{\linewidth}{!}{%
\begin{tabular}{|c|c|ccccccc|}
\hline
    \multirow{2}{*}{\textbf{In-distriution Dataset name}} & \multirow{2}{*}{MAE} & \multicolumn{7}{c|}{\textbf{OOD Dataset label}} \\
    & & DS1 & DS2 & DS3 & DS4 & DS5 & DS6 & DS7 \\ \hline
ArticW. (Motion) & 0.025  & CharacterT. & EigenW. & PenD. & $\emptyset$  & $\emptyset$  & $\emptyset$  & $\emptyset$ \\\hline   
EigenW. (Motion) & 0.000   & ArticW. & CharacterT. & PenD. & $\emptyset$  & $\emptyset$  & $\emptyset$  & $\emptyset$ \\\hline  
PenD. (Motion) & 0.001   & ArticW. & CharacterT. & EigenW. & $\emptyset$  & $\emptyset$  & $\emptyset$  & $\emptyset$ \\\hline  
AtrialF. (ECG) & 0.005   & StandW. & $\emptyset$  & $\emptyset$  & $\emptyset$  & $\emptyset$  & $\emptyset$  & $\emptyset$ \\\hline  
StandW. (ECG)  & 0.012  & AtrialF. & $\emptyset$  & $\emptyset$  & $\emptyset$  & $\emptyset$  & $\emptyset$  & $\emptyset$ \\\hline  
BasicM. (HAR)  & 0.024  & Cricket & Epilepsy & Handw. & Libras & NATOPS & RacketS. & UWaveG.\\\hline  
Cricket (HAR)  & 0.010  & BasicM. & Epilepsy & Handw. & Libras & NATOPS & RacketS. & UWaveG.\\\hline  
Epilepsy (HAR) & 0.030   & BasicM. & Cricket & Handw. & Libras & NATOPS & RacketS. & UWaveG.\\\hline  
Handw. (HAR) & 0.006   & BasicM. & Cricket & Epilepsy & Libras & NATOPS & RacketS. & UWaveG.\\\hline  
Libras (HAR) & 0.003   & BasicM. & Cricket & Epilepsy & Handw. & NATOPS & RacketS. & UWaveG.\\\hline  
NATOPS (HAR)  & 0.046  & BasicM. & Cricket & Epilepsy & Handw. & Libras & RacketS. & UWaveG.\\\hline  
RacketS. (HAR)  & 0.026  & BasicM. & Cricket & Epilepsy & Handw. & Libras & NATOPS & UWaveG.\\\hline  
UWaveG. (HAR)  & 0.015  & BasicM. & Cricket & Epilepsy & Handw. & Libras & NATOPS & RacketS.\\\hline  
EthanolC. (Other) & 0.001   & ER. & LSST & PEMS-SF & $\emptyset$  & $\emptyset$  & $\emptyset$  & $\emptyset$ \\\hline  
ER. (Other)  & 0.044  & EthanolC. & LSST & PEMS-SF & $\emptyset$  & $\emptyset$  & $\emptyset$  & $\emptyset$ \\\hline  
LSST (Other) & 0.008   & EthanolC. & ER. & PEMS-SF & $\emptyset$  & $\emptyset$  & $\emptyset$  & $\emptyset$ \\\hline  
PEMS-SF (Other) & 0.525   & EthanolC. & ER. & LSST & $\emptyset$  & $\emptyset$  & $\emptyset$  & $\emptyset$ \\\hline  
FingerM. (EEG) & 0.048   & HandM. & MotorI. & SelfR1. & SelfR2. & $\emptyset$  & $\emptyset$  & $\emptyset$ \\\hline  
HandM. (EEG)  & 0.006  & FingerM. & MotorI. & SelfR1. & SelfR2. & $\emptyset$  & $\emptyset$  & $\emptyset$ \\\hline  
MotorI. (EEG) & 0.543   & FingerM. & HandM. & SelfR1. & SelfR2. & $\emptyset$  & $\emptyset$  & $\emptyset$ \\\hline  
SelfR1. (EEG) & 0.009   & FingerM. & HandM. & MotorI. & SelfR2. & $\emptyset$  & $\emptyset$  & $\emptyset$ \\\hline  
SelfR2. (EEG)  & 0.012  & FingerM. & HandM. & MotorI. & SelfR1. & $\emptyset$  & $\emptyset$  & $\emptyset$ \\\hline  
Heartbeat (Audio)  & 0.011  & JapaneseV. & SpokenA. & $\emptyset$  & $\emptyset$  & $\emptyset$  & $\emptyset$  & $\emptyset$ \\\hline   

\end{tabular}
}
\end{table*}

\vspace{1.0ex}

\noindent \textbf{Evaluation metrics.} 
We employ the following two standard metrics in our experimental evaluation. {\bf 1) AUROC score:} The area under the receiver operating characteristic curve is a threshold-independent metric. 
This metric (higher is better) is equal to 1.0 for a perfect detector and 0.5 for a random detector. {\bf 2) F1 score:}  It is the harmonic mean of precision and recall. Due the threshold dependence of F1 score, we propose to use the highest F1 score obtained with a variable threshold. This score has a maximum of 1.0 in the case of a  perfect precision and recall. 

\vspace{1.0ex}

\noindent \textbf{Configuration of algorithms.} We employ a 1D-CNN architecture for the CVAE models required for seasonal ratio scoring (SR) method. We consider a naive baseline where the CVAE is trained on the ID data and the likelihood (LL) is used to detect OOD samples. We consider a variant of SR scoring (SR$_a$) that works on the aligned time-series data using the method explained in Section 4.3. We evaluate both SR and SR$_a$ against state-of-the-art baselines and employ their publicly available code: Out-of-Distribution Images in Neural networks (ODIN) \citep{liang2017enhancing} and Gram Matrices (GM) \citep{sastry20gram} that have been shown to outperform most of the existing baselines; recently proposed Likelihood Regret (LR) score \citep{xiao2020regret}; adaptation of a very recent time-series AD method referred to as Deep generative model with hierarchical latent (HL) \citep{challu2022deep} that does not require labeled anomalies for training purposes. We chose HL as the main baseline to represent time-series AD under OOD setting as it is the state-of-the-art  time-series AD algorithm. HL for time-series was shown \citep{challu2022deep} to outperform nearest-neighbor based methods, LSTM-based methods, and other methods \citep{blazquez2021review,braei2020anomaly} in various AD settings.

\begin{itemize}
    \item \textbf{Choice of architecture:} We have experimented with 3 different types of CVAE architecture to decide on the most suitable one for our OOD experiments. We evaluated 1) fully connected, 2) convolutional, and 3) LSTM based architectures using the reconstruction error as the performance metric. 
    We have observed that fully connected networks generally suffer from poor reconstruction performance, especially on high-dimensional data. We have also observed that LSTM's runtime during training and inference is relatively longer than the other architectures. However, CNN-based CVAEs delivered both a good reconstruction performance and fast runtime.

    \item \textbf{1D-CNN CVAE details:} To evaluate the effectiveness of the proposed seasonal ratio (SR) score, we employed a CVAE that is based on 1D-CNN layers. The encoder of the CVAE is composed of 1) A minmax normalization layer, 2) A series of 1D-CNN layers, and 3) A fully-connected layer. At the end of the encoder, the parameters $\mu_{\text{CVAE}}$ and $\sigma_{\text{CVAE}}$ are computed to estimate the posterior distribution. A random sample is then generated from this distribution and passed on to the CVAE decoder along with the class label. The decoder of the CVAE is composed of 1) A fully-connected layer, 2) A series of transposed convolutional layer, and 3) A denormalization layer.

    \item \textbf{CVAE Training:} We use the standard training, validation, and testing split on the benchmark datasets to train both CVAEs $\mathcal{M}_x$ and $\mathcal{M}_r$. Both CVAEs are trained to maximize the ELBO on the conditional log-likelihood defined in Section 3 using Adam optimizer with a learning rate of $10^{-4}$. We employ a maximum number of training iterations equal to $500$. To ensure the reliability of the performance of the proposed CVAEs, we report in Table \ref{tab:dslabel} the test reconstruction error of the trained CVAE on ID data using Mean Absolute Error (MAE). We observe clearly that the proposed CVAE is able to learn well the ID space as the reconstruction error is relatively low.  To compute the semantic patterns and remainders for in-distribution examples for training $\mathcal{M}_x$ and $\mathcal{M}_r$, we use the  STLdecompose\footnote{https://github.com/jrmontag/STLDecompose.git} python package.

    \item {\bf Implementation of the baselines:} The baseline methods for ODIN\footnote{https://github.com/facebookresearch/odin.git}, GM\footnote{https://github.com/VectorInstitute/gram-ood-detection.git}, HL\footnote{https://github.com/cchallu/dghl.git} and LR\footnote{https://github.com/XavierXiao/Likelihood-Regret.git} were implemented using their respective publicly available code with the recommended settings. To employ ODIN and GM, we have trained two different DNN models: a 1D-CNN and an LSTM for classification tasks with different settings. We report the average performance of the baseline OOD detectors in our experimental setting. To repurpose HL method from the AD setting to OOD setting, we have serialized the training data and use it during the training of the generator. For OOD detection at inference time, we serialize both the test ID data and OOD data and shuffle them. By providing the window size equal to the time-steps dimension of the original in-distribution inputs, we execute HL anomaly detection algorithm and report every anomaly as an OOD sample. We employed the default parameters of the generator. As recommended by the authors, we use a hierarchical level equal to 4 and 500 iterations for training and inference. We lower the learning rate to $10^{-6}$ to prevent the exploding gradient that occurred using the default $10^{-3}$ value. For a fair comparison, the VAE for Likelihood Regret (LR) has the same architecture as the CVAE used to estimate the seasonal ratio (SR) and the naive LL score.
\end{itemize}

\subsection{Results and Discussion}

\vspace{1.0ex}

\noindent \textbf{Reconstruction error of DGMs.} Table \ref{tab:OODrecons} shows the test reconstruction error of the trained CVAE on ID data using Mean Absolute Error (MAE). We clearly observe that CVAE model is able to learn the ID space as the reconstruction error is relatively low.  Table \ref{tab:OODrecons} shows analogous results for the same CVAE on some OOD data. We observe that DGMs perform well on OOD samples regardless of the different semantics of both ID and OOD data. The pre-trained CVAEs performed well on the OOD AWR dataset with a reconstruction error $\le 0.1$. For the OOD FingerMovement (Fmv) dataset, only two out of the six CVAEs exhibited an intuitive high reconstruction error.
\begin{table}[!h]
\centering
\caption{Average reconstruction error of CVAE is small on both ID and OOD data. The variance is  $\le 0.001$.}
\label{tab:OODrecons}
\begin{tabular}{cc||cccccc}
\multicolumn{2}{c||}{\textbf{ID Train}} & AWR & SWJ & Ckt & HMD & Hbt & ERg \\ \hline
\multicolumn{2}{c||}{\textbf{ID Test}} & 0.025 & 0.012 & 0.010 & 0.118 & 0.011 & 0.045 \\ \hline
\multirow{2}{*}{\textbf{OOD}} & AWR & $\emptyset$ & 0.018 & 0.035 & 0.039 & 0.002 & 0.137 \\
& Fmv & 1.658 & 0.146 & 0.146 & 0.039 & 0.071 & 6.132 
\end{tabular}
\end{table}

\vspace{1.0ex}

\noindent \textbf{OOD detection via pre-trained classifier and DGMs.} Our first hypothesis is that pre-trained DNN classifiers are not well-suited for OOD detection. To test this hypothesis, we train two DNN models: a 1D-CNN and an RNN classifier. We use these models for OOD detection using the ODIN and GM baselines. Table \ref{tab:OODpretrained} shows that AUROC is low on all datasets. For datasets such as HMD and SWJ, the AUROC score does not exceed 0.6 for any experimental setting. The accuracy of DNNs for time-series classification is not as high as those for the image domain for the reasons explained earlier. Hence, we believe that this uncertainty of DNNs causes the baselines ODIN and GM to fail in OOD detection.
\begin{table}[!h]
\setlength\extrarowheight{2pt}
\caption{Average AUROC results for ODIN and GMM.}
\label{tab:OODpretrained}
\begin{tabular}{cc||ccccccc}
 & & AWR & SWJ & Ckt &  HMD & Hbt & ERg \\ \hline
\multirow{2}{*}{\shortstack{\textbf{In-}\\ \textbf{Domain}}}& ODIN& 0.65$\pm 0.03$ & 0.54$\pm 0.01$ & 0.65$\pm 0.03$&  0.55$\pm 0.01$ & 0.71$\pm 0.03$ & 0.70$\pm 0.03$\\ 
 & GM  & 0.70$\pm 0.03$ & 0.58$\pm 0.01$ & 0.64$\pm 0.03$&  0.58$\pm 0.02$ & 0.80$\pm 0.01$ & 0.75$\pm 0.03$  \\  \hline
 \multirow{2}{*}{\shortstack{\textbf{Cross-}\\\textbf{Domain}}}&ODIN& 0.55$\pm 0.01$ & 0.54$\pm 0.01$ & 0.55$\pm 0.01$&  0.50 & 0.70$\pm 0.01$ & 0.70$\pm 0.01$ \\ 
 & GM & 0.56$\pm 0.01$ & 0.54$\pm 0.01$ & 0.65$\pm 0.03$&  0.55$\pm 0.01$ & 0.75$\pm 0.03$ & 0.78$\pm 0.02$  
\end{tabular}
\end{table}
Our second hypothesis is that DGMs assign a high likelihood for OOD samples is also applicable to time-series. While results in Table \ref{tab:OODrecons} corroborated this hypothesis, we provide the use of a pre-trained CVAE for OOD detection (LL) in Table \ref{tab:auroc}. We observe that AUROC score of LL does not outperform any of the other baselines. Hence, a new scoring method is necessary for CVAE-based OOD detection.

\begin{figure}[t]
\centering
    \begin{minipage}{.42\linewidth}
        \centering
        \includegraphics[width=\linewidth]{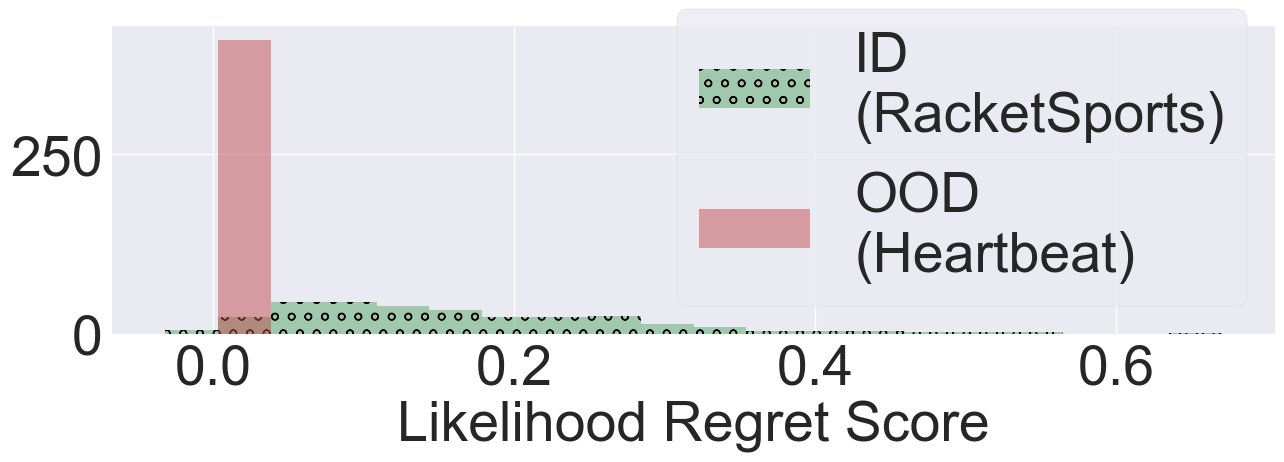}
    \end{minipage}%
    \begin{minipage}{.42\linewidth}
        \centering
        \includegraphics[width=\linewidth]{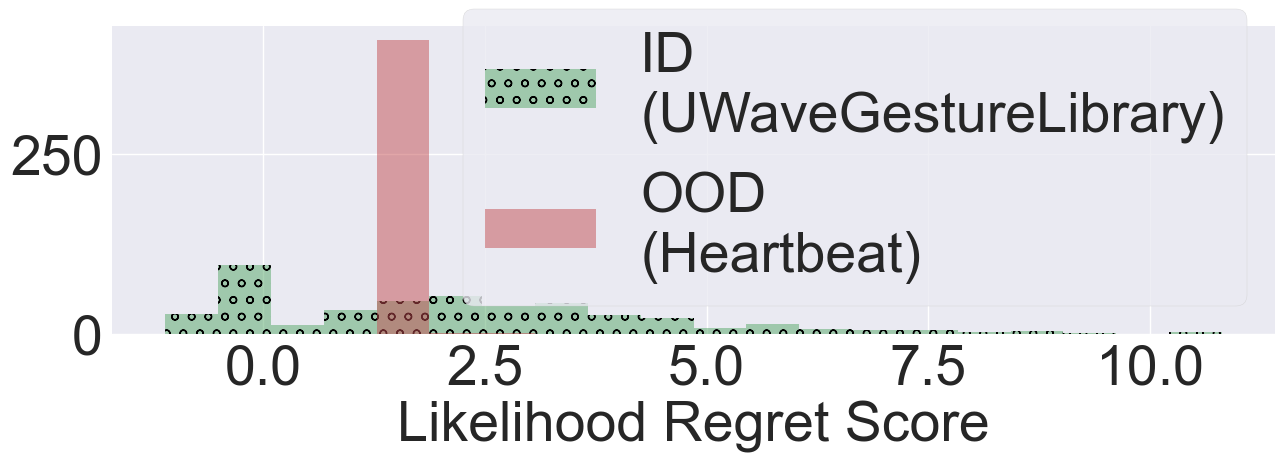}
    \end{minipage}
    \begin{minipage}{.42\linewidth}
        \centering
        \includegraphics[width=\linewidth]{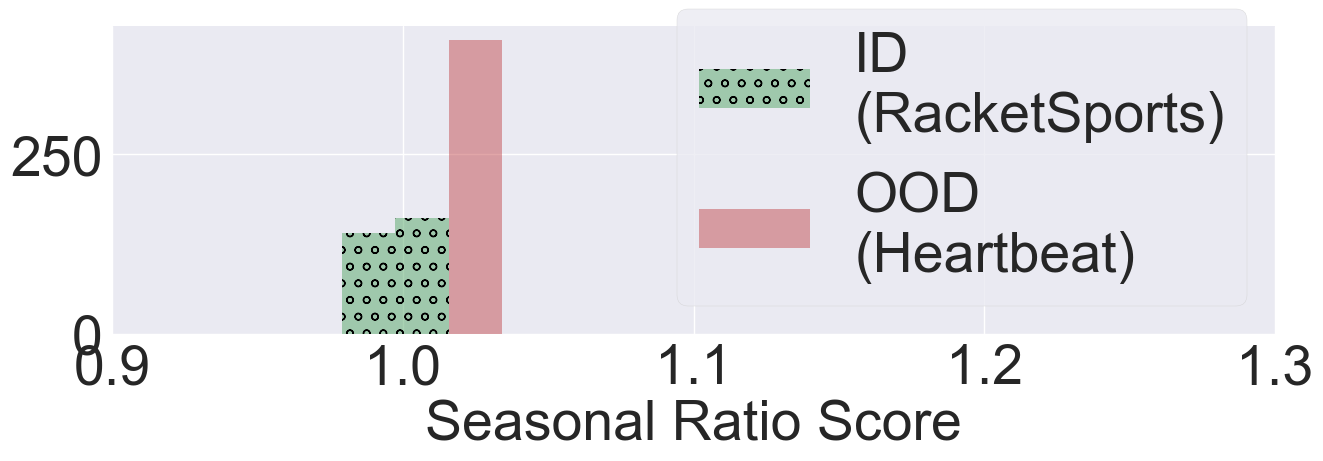}
    \end{minipage}%
    \begin{minipage}{.42\linewidth}
        \centering
        \includegraphics[width=\linewidth]{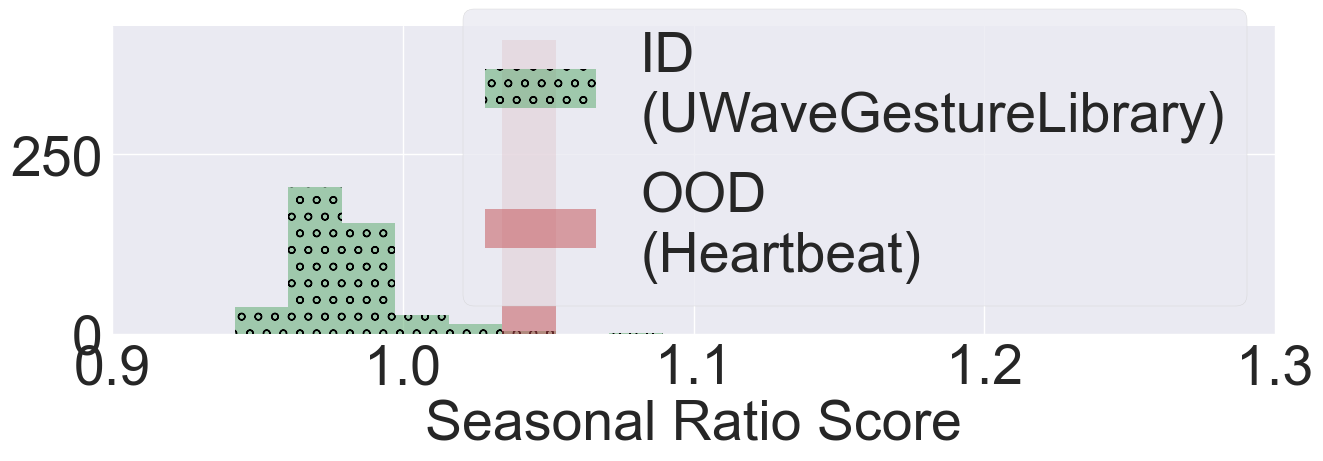}
    \end{minipage}
    \caption{Histogram showing the non-separability of ID and OOD LR scores (Top row) and the separability using the seasonal ratio method on real-world time-series data (Bottom row).}
    \label{fig:histoverlap}
\end{figure}

\vspace{1.0ex}

\noindent \textbf{Random Noise as OOD.} An existing experimental setting for OOD detection tasks is to detect random noise. For this setting, we generate random noise as an input sampled from a Gaussian distribution or a Uniform distribution. Table \ref{tab:noiseOOD} shows the LR baseline performance on detecting the random noise as OOD examples. We observe in Table \ref{tab:noiseOOD} that LR has an excellent performance on this task. This is explainable as time-series noise does not necessarily obey time-series characteristics. Hence, the existing baselines can perform strongly on the OOD examples. We motivate our seasonal ratio scoring approach for OOD detection based on real-world examples. We have shown in the main paper that existing baselines have poor performance in detecting real-world OOD examples, whereas SR has a significantly better performance.

\begin{table}[!h]
\centering
\caption{Average performance of LR on OOD examples sampled from Gaussian/Uniform distribution.}
\label{tab:noiseOOD}
\begin{tabular}{c||cccccc}
 & AWR & SWJ & Ckt & HMD & Hbt & ERg \\ \hline
 \textbf{LR}& 1.00 & 1.00 & 1.00 & 1.00 & 0.95 & 1.00 \\ 
\end{tabular}
\end{table}
\begin{table}[!h]
\centering
\caption{Results for the validity of Assumption \ref{asp:main}. Average distance (MAE and DTW measures) between the semantic pattern from STL $S_y$ and time-series example $x$ with label $y$ from the testing data (with a negligible variance$\le 0.001$).}
\label{tab:patternerr}
\begin{tabular}{c||cccccc}
 & AWR & SWJ & Ckt & HMD & Hbt & ERg \\ \hline
\textbf{MAE} & 0.047 & 0.031 & 0.029 & 0.078 & 0.002 & 0.086  \\ \hline
\textbf{DTW} & 0.039 & 0.018 & 0.022 & 0.068 & 0.002 & 0.069 
\end{tabular}
\end{table}

\vspace{1.0ex}

\noindent {\bf Results for SR score.} The effectiveness of SR score depends on the validity of Assumption \ref{asp:main}. Table \ref{tab:patternerr} shows both MAE and DTW measure between semantic pattern $S_y$ from STL and different time-series examples of the same class $y$. We observe that the average difference measure is low. These results strongly demonstrate that  Assumption \ref{asp:main} holds empirically.
For qualitative results, Figure \ref{fig:histoverlap} shows the performance of SR in contrast to the performance of Likelihood Regret (LR) shown in Figure \ref{fig:histoverlap}. This illustration shows that SRS provides significantly better OOD separability.

\begin{table*}[b]
\setlength\extrarowheight{2.5pt}
\caption{AUROC results for the baselines, SR, and SR with time-series alignment (SR$_a$) on different datasets for both in-domain and cross-domain OOD setting. {\bf DSi} is a label given to the  dataset used as real-world OOD data compared to the ID dataset in the first cell of every column. {\bf Di} is a label given to the domain used to define the OOD distribution. $\emptyset$ denotes a non-existent setting, e.g., AWR's corresponding domain has only three datasets. The last two rows show the percentage of total experiments where SR$_a$ outperforms the baseline methods, and SR$_a$ improves over $SR$ performance respectively.}
\label{tab:auroc}
\resizebox{\linewidth}{!}{%
\begin{tabular}{|cc|ccccccc||cccccc|}
\cline{3-15}
 \multicolumn{2}{c|}{ } &  \multicolumn{7}{c||}{\textbf{In-domain OOD}} &  \multicolumn{6}{c|}{\textbf{Cross-domain OOD}}  \\ \cline{3-15}
 \multicolumn{2}{c|}{ } & DS1 & DS2 & DS3 & DS4 & DS5 & DS6 & DS7 & D1 & D2 & D3 & D4 & D5 & D6 \\ \hline
\multirow{3}{*}{\shortstack{AWR\\(Motion)}} 
    & LL & 0.80 & 0.85 & 0.54 & $\emptyset$ & $\emptyset$ & $\emptyset$ & $\emptyset$ & $\emptyset$ & 0.81 & 0.80 & 0.57  & 0.59 & 0.81 \\
    & HL & \emph{0.50} & 0.96 & 0.94 & $\emptyset$ & $\emptyset$ & $\emptyset$ & $\emptyset$ & $\emptyset$ & \emph{0.50} & 0.75 & 0.98  & \emph{0.50} & 0.56 \\ 
    & LR & 0.90 & 0.95 & 0.66 & $\emptyset$ & $\emptyset$ & $\emptyset$ & $\emptyset$ & $\emptyset$ & 0.61 & 0.84 & 0.56  & 0.72 & 0.61 \\
    & SR & 0.90 & 0.97 & 0.95 & $\emptyset$ & $\emptyset$ & $\emptyset$ & $\emptyset$ & $\emptyset$ & 1.00 & 0.97 & 1.00  & 1.00 & 1.00 \\ 
    & SR$_a$ & 0.90 &  \textbf{0.97}&  \textbf{0.95}& $\emptyset$ & $\emptyset$ & $\emptyset$ & $\emptyset$ & $\emptyset$ &  \textbf{1.00}&   \textbf{1.00}&   \textbf{1.00}&   \textbf{1.00} &  \textbf{1.00} \\ \hline
\multirow{3}{*}{\shortstack{SWJ\\(ECG)}} 
    & LL & 0.55 & $\emptyset$ & $\emptyset$ & $\emptyset$ & $\emptyset$ & $\emptyset$ & $\emptyset$ & 0.61 & $\emptyset$ & 0.51 & 1.00  &  0.77 & 0.52 \\ 
    & HL & \emph{0.50} & $\emptyset$ & $\emptyset$ & $\emptyset$ & $\emptyset$ & $\emptyset$ & $\emptyset$ & \emph{0.50} & $\emptyset$ & \emph{0.50} & \emph{0.50}  & \emph{0.50} & \emph{0.50} \\ 
    & LR & \textbf{0.97} & $\emptyset$ & $\emptyset$ & $\emptyset$ & $\emptyset$ & $\emptyset$ & $\emptyset$ & 0.64 & $\emptyset$ & \emph{0.50} & 1.00  &  0.67 &  \textbf{0.99} \\ 
    & SR &  0.65& $\emptyset$ & $\emptyset$ & $\emptyset$ & $\emptyset$ & $\emptyset$ & $\emptyset$ & 0.70 & $\emptyset$ &  0.61 & 1.00  &  0.96 & 0.61 \\  
    & SR$_a$ & 0.67 & $\emptyset$ & $\emptyset$ & $\emptyset$ & $\emptyset$ & $\emptyset$ & $\emptyset$ &  \textbf{0.70} & $\emptyset$ &  \textbf{0.61} & 1.00  &   \textbf{0.96} &  0.61 \\ \hline
\multirow{3}{*}{\shortstack{Ckt\\(HAR}} 
    & LL & 0.89  & 0.85 & 0.84 & 0.81 & 0.82 & 0.91 & 0.79 & 0.79 & 0.82 & $\emptyset$ & 0.95 & 0.90 & 0.81 \\ 
    & HL & 0.97 & \emph{0.50} & 0.99 & \emph{0.50} & \emph{0.50} & 0.51 & \emph{0.50} & \emph{0.50} & \emph{0.50} & $\emptyset$ & 0.52 & 0.94 & \emph{0.50} \\
    & LR & 0.81 & 0.75 & 0.74 & 0.71 & 0.74 & 1.00 & 0.78 & 0.77 & 0.72 & $\emptyset$ & 0.95 & 0.88 & 0.71 \\ 
    & SR & 0.99 & 0.98 & 0.99 & 0.99 & 0.99 & 0.98 & 0.98 & 0.98 & 0.99 & $\emptyset$ & 0.99 & 0.98 & 0.98 \\ 
    & SR$_a$ &  \textbf{0.99} & \textbf{0.99} &  \textbf{1.00} &  \textbf{1.00} & \textbf{1.00} & 1.00& \textbf{0.98} & \textbf{0.98} & \textbf{0.99} & $\emptyset$ &  \textbf{1.00}  &  \textbf{1.00} &  \textbf{1.00} \\   \hline
\multirow{3}{*}{\shortstack{HMD\\(EEG)}} 
    & LL & 0.88 & 0.88 & 0.89 & 0.89 & $\emptyset$ & $\emptyset$ & $\emptyset$ & 0.89 & 0.80 & 0.87 & $\emptyset$ & 0.90 & 0.91 \\ 
    & HL & \textbf{0.93} & 0.57 & 0.78 & 0.87 & $\emptyset$ & $\emptyset$ & $\emptyset$ & \textbf{0.97} & \emph{0.50} & \textbf{0.98} & $\emptyset$ & 0.58 & 0.66 \\ 
    & LR & 0.68 & 0.68 & 0.68 & 0.68 & $\emptyset$ & $\emptyset$ & $\emptyset$ & 0.68 & 0.68 & 0.68 & $\emptyset$ & 0.80 & 0.68 \\ 
    & SR & 0.75 & 0.75 & 0.75 & 0.75 & $\emptyset$ & $\emptyset$ & $\emptyset$ &  0.75 & 0.75 & 0.75 & $\emptyset$ & 0.83 & 0.75 \\ 
    & SR$_a$ &  0.90&  \textbf{0.90} &  \textbf{0.90} &  \textbf{0.90} & $\emptyset$ & $\emptyset$ & $\emptyset$ &   0.75&  \textbf{0.84} &  0.89& $\emptyset$ &  \textbf{0.97} &  0.91 \\   \hline
\multirow{3}{*}{\shortstack{Hbt\\(Audio)}} 
    & LL & 1.00 & 1.00 & $\emptyset$ & $\emptyset$ & $\emptyset$ & $\emptyset$ & $\emptyset$ & 0.90 & 0.95 & 0.94  & 0.85 & $\emptyset$ & 0.98 \\ 
    & HL & \emph{0.50} & \emph{0.50} & $\emptyset$ & $\emptyset$ & $\emptyset$ & $\emptyset$ & $\emptyset$ & 0.96 & \emph{0.50} & 0.78  & 0.62 & $\emptyset$ & 0.82 \\ 
    & LR & 1.00 & 1.00 & $\emptyset$ & $\emptyset$ & $\emptyset$ & $\emptyset$ & $\emptyset$ & 0.93 & 0.94 & 0.94  & 0.75 & $\emptyset$ & 0.98 \\ 
    & SR & 1.00 & 1.00 & $\emptyset$ & $\emptyset$ & $\emptyset$ & $\emptyset$ & $\emptyset$ & 0.96 & 0.97 & 0.94  & 1.00 & $\emptyset$ & 1.00 \\ 
    & SR$_a$ & 1.00 & 1.00 & $\emptyset$ & $\emptyset$ & $\emptyset$ & $\emptyset$ & $\emptyset$ &  0.96&  \textbf{0.97}& 0.94  &   \textbf{1.00}& $\emptyset$ &   \textbf{1.00} \\  \hline
\multirow{3}{*}{\shortstack{ERg\\(Other)}} 
    & LL & 0.83 & 0.77 & 0.75 & $\emptyset$ & $\emptyset$ & $\emptyset$ & $\emptyset$ & 0.88 & 0.86 & 0.82 & 0.77 & 0.76  & $\emptyset$\\ 
    & HL & \emph{0.50} & \emph{0.50} & \emph{0.50} & $\emptyset$ & $\emptyset$ & $\emptyset$ & $\emptyset$ & \emph{0.50} & \emph{0.50} & \emph{0.50} & \emph{0.50} & \emph{0.50}  & $\emptyset$\\ 
    & LR & 0.83 & 0.72 & 0.78 & $\emptyset$ & $\emptyset$ & $\emptyset$ & $\emptyset$ & 0.88 & 0.78 & 0.81 & 0.71 & 0.78  & $\emptyset$\\ 
    & SR & 1.00 & 0.98 & 0.99 & $\emptyset$ & $\emptyset$ & $\emptyset$ & $\emptyset$ & 0.89 & 0.99 & 0.94 & 0.99 & 0.99  & $\emptyset$\\ 
    & SR$_a$ & \textbf{ 1.00}&  \textbf{1.00}&  \textbf{1.00}& $\emptyset$ & $\emptyset$ & $\emptyset$ & $\emptyset$ &  \textbf{0.95}&  \textbf{1.00}&  \textbf{0.95}&  \textbf{1.00}&  \textbf{1.00}& $\emptyset$ \\   \hline
    \multicolumn{2}{|c|}{Absolute } &
    \multicolumn{2}{c}{LL 00.0\%}  & 
    \multicolumn{2}{c}{HL 05.0\%}  &
    \multicolumn{3}{c||}{LR 05.0\%} & 
    \multicolumn{2}{c}{LL 0.00\%}  &
    \multicolumn{2}{c}{HL 06.7\%}  &
    \multicolumn{2}{c|}{LR 03.30\%} 
    \\ 
   \multicolumn{2}{|c|}{ Wins (\%)} &
   \multicolumn{3}{c}{Ties 20.0\%}  &
   \multicolumn{4}{c||}{\textbf{SR$_a$ 70.0\%}} &
   \multicolumn{3}{c}{Ties 16.7\%}  &
   \multicolumn{3}{c|}{\textbf{SR$_a$ 73.3\%}}  
    \\ \hline
\multicolumn{2}{|c|}{$SR_a$ Improvement (\%)} & \multicolumn{7}{c||}{55.0\%} & \multicolumn{6}{c|}{40.0\%} \\ 
    \hline
\end{tabular}
}
\end{table*}

\vspace{1.0ex}

\noindent {\bf SR score vs. Baselines.} Table \ref{tab:auroc} shows the OOD results for SRS and baseline methods. For a fair comparison, we use the same architecture for VAEs computing the LL, LR, and SR scores. We make the following observations. {\bf 1)} The naive LL method fails to outperform any other approach, which demonstrates that DGMs are not reliable on their own as they produce high likelihood for OOD samples. {\bf 2)} The time-series anomaly detection method HL fails drastically for various OOD settings as reflected by the poor AUROC score of 0.5. This demonstrates that AD methods are not appropriate for OOD detection in the multi-class setting. {\bf 3)} SR score outperforms LR score in identifying OOD examples on 80\% of the total experiments. This means that the improvement is due to a better scoring function. {\bf 4)} For the in-domain OOD setting, AUROC score of LR is always lower than SRS. 3) For the cross-domain setting, SRS outperforms LR in all cases except one experiment on a single dataset SWJ. 4) LR and SRS have the same performance in 20\% of the total experiments. Therefore, we conclude that SRS is better than LR in terms of OOD performance and execution time (LR requires new training for every single testing input unlike SR).

\vspace{1.0ex}

\noindent \textbf{Alignment improves the accuracy of SR score.}
Our hypothesis is that extraction of an accurate semantic component using STL results in improved OOD detection accuracy. To test this hypothesis, we compare SR and SR$_a$ (SR with aligned time-series data). Table \ref{tab:auroc} shows the AUROC scores of SR and SR$_a$. SR$_a$ improves the performance of SR for around 50\% of the overall experiments. For example, on HMD dataset, we observe that SR$_a$ enhances the performance of SR by an average of 15\% under the in-domain OOD setting. 
These results strongly corroborate our hypothesis that alignment improves OOD performance.

\vspace{1.0ex}

\noindent \textbf{SR performance using F1-score.}
In addition to the AUROC score, we employ F1 score to assess the effectiveness of SR score in detecting OOD. Table \ref{tab:f1sc} provides the results comparing SR score and LR score. Like AUROC score evaluation, we make similar observations on F1 score. {\bf 1)} SR score outperforms LR score in identifying OOD examples on 60\% of the total experiments. This means improvement is due to better scoring function. {\bf 2)} For the in-domain OOD setting, F1 score of LR is mostly lower than SR. 3) For the cross-domain setting, SR outperforms LR in 66\% of the cases. Hence, we conclude that SR is better than LR in terms of OOD performance measured as F1 metric.

\begin{table*}[!h]
\setlength\extrarowheight{2pt}
\caption{F1 metric results of LR, $SR_a$ on the different datasets for both In-Domain and Cross-Domain setting. The last two rows show the percentage of datasets where $SR_a$ is out-performing the $LR$ score.}
\label{tab:f1sc}
\resizebox{\linewidth}{!}{%
\begin{tabular}{|cc|ccccccc||cccccc|}
\cline{3-15}
 \multicolumn{2}{c|}{ } &  \multicolumn{7}{c||}{\textbf{In-Domain OOD}} &  \multicolumn{6}{c|}{\textbf{Cross-Domain OOD}}  \\ \cline{3-15}
 \multicolumn{2}{c|}{ } & DS1 & DS2 & DS3 & DS4 & DS5 & DS6 & DS7 & D1 & D2 & D3 & D4 & D5 & D6 \\ \hline
\multirow{2}{*}{\shortstack{AWR\\(Motion)}} 
    & LR & 0.58 & \textbf{0.99} & 0.80 & $\emptyset$ & $\emptyset$ & $\emptyset$ & $\emptyset$ & $\emptyset$ & 0.53 & 0.79 & 0.61  & 0.44 & 0.67 \\
    & SR$_a$ & \textbf{0.69 }& 0.89 & \textbf{0.97} & $\emptyset$ & $\emptyset$ & $\emptyset$ & $\emptyset$ & $\emptyset$ & \textbf{0.97} & \textbf{0.85} & \textbf{1.00} & \textbf{0.99} & \textbf{1.00} \\ \hline
\multirow{2}{*}{\shortstack{SWJ\\(ECG)}} 
    & LR & 0.69 & $\emptyset$ & $\emptyset$ & $\emptyset$ & $\emptyset$ & $\emptyset$ & $\emptyset$ & 0.98 & $\emptyset$ & 0.82 & 1.00  &  0.97 & 0.97 \\ 
    & SR$_a$ & 0.69 & $\emptyset$ & $\emptyset$ & $\emptyset$ & $\emptyset$ & $\emptyset$ & $\emptyset$ & 0.98 & $\emptyset$ & \textbf{0.86} & 1.00  &  0.97 & \textbf{0.98} \\ \hline
\multirow{2}{*}{\shortstack{Ckt\\(HAR)}} 
    & LR & 0.70 & 0.90 & 0.97 & 0.92 & 0.93 & 0.91 & 0.95 & 0.96 & 0.48 & $\emptyset$ & \textbf{1.00} & 0.94 & 0.93\\ 
    & SR$_a$ & \textbf{0.98} & \textbf{0.96} & \textbf{0.99} & \textbf{0.98} & \textbf{0.99} & \textbf{0.99} & \textbf{0.98} & \textbf{0.98} & \textbf{0.94}  & $\emptyset$ & 0.99  & \textbf{0.97} & \textbf{1.00} \\   \hline
\multirow{2}{*}{\shortstack{HMD\\(EEG)}} 
    & LR & \textbf{0.82} & \textbf{0.81} & 0.84 & 0.81 & $\emptyset$ & $\emptyset$ & $\emptyset$ & 0.84 & 0.45 & 0.68 & $\emptyset$ & 0.80 & 0.82 \\ 
    & SR$_a$ & 0.80 & 0.75 & 0.84 & 0.81 & $\emptyset$ & $\emptyset$ & $\emptyset$ & \textbf{0.94} &\textbf{ 0.65} & 0.68 & $\emptyset$ & \textbf{0.90} & \textbf{0.92} \\  \hline
\multirow{2}{*}{\shortstack{Hbt\\(Audio)}} 
    & LR & 1.00 & 1.00 & $\emptyset$ & $\emptyset$ & $\emptyset$ & $\emptyset$ & $\emptyset$ & 0.93 & 0.88 & 0.94  & 0.75 & $\emptyset$ & 0.98 \\ 
    & SR$_a$ & 1.00 & 1.00 & $\emptyset$ & $\emptyset$ & $\emptyset$ & $\emptyset$ & $\emptyset$ & 0.93 & 0.88 & 0.94  & \textbf{1.00} & $\emptyset$ & 0.98 \\  \hline
\multirow{2}{*}{\shortstack{ERg\\(Other)}} 
    & LR & 0.44 & 0.85 & 0.76 & $\emptyset$ & $\emptyset$ & $\emptyset$ & $\emptyset$ & \textbf{0.88} & 0.90 & 0.79 & 0.83 & 0.88 & $\emptyset$ \\ 
    & SR$_a$ & \textbf{0.99} & \textbf{0.99} & \textbf{0.95} & $\emptyset$ & $\emptyset$ & $\emptyset$ & $\emptyset$ & 0.87 & \textbf{0.94} & \textbf{0.88} & \textbf{0.98} & \textbf{1.00} & $\emptyset$ \\   \hline
\multirow{2}{*}{Wins(\%)} 
    & LR & \multicolumn{7}{c||}{15.0\%} & \multicolumn{6}{c|}{6.7\%}  \\ 
    & Ties & \multicolumn{7}{c||}{25.0\%} & \multicolumn{6}{c|}{26.7\%}  \\ 
    & SR$_a$ & \multicolumn{7}{c||}{60.0\%} & \multicolumn{6}{c|}{66.6\%}  \\  
    \hline
\end{tabular}
}
\end{table*}

\vspace{1.0ex}

\noindent \textbf{AUROC performance of SR scoring on the full multivariate UCR dataset.}
For the sake of completeness, Table \ref{tab:auroc2} provides additional results for the performance of SR on all the UCR multi-variate datasets in terms of the AUROC score. These results demonstrate that the proposed SR scoring approach is general and highly effective for all time-series datasets.

\begin{table*}[t]
    \centering
\caption{AUROC results for LR, and SR$_a$ on different datasets for both in-domain and cross-domain OOD setting. The last two rows shows respectively the percentage of datasets where $SR_a$ is out-performing the $LR$ score.}
\label{tab:auroc2}
\resizebox{.85\linewidth}{!}{%
\begin{tabular}{|cc|ccccccc||cccccc|}
\cline{3-15}
 \multicolumn{2}{c|}{ } &  \multicolumn{7}{c||}{\textbf{In-domain OOD}} &  \multicolumn{6}{c|}{\textbf{Cross-domain OOD}}  \\ \cline{3-15}
 \multicolumn{2}{c|}{ } & DS1 & DS2 & DS3 & DS4 & DS5 & DS6 & DS7 & D1 & D2 & D3 & D4 & D5 & D6 \\ \hline
 
\multirow{3}{*}{\shortstack{ArticularyW.}} 
    & LR & 0.90 & 0.95 & 0.66 & $\emptyset$ & $\emptyset$ & $\emptyset$ & $\emptyset$ & $\emptyset$ & 0.61 & 0.84 & 0.56  & 0.72 & 0.61 \\
    & SR$_a$ & 0.90 & \textbf{0.97} & \textbf{0.95} & $\emptyset$ & $\emptyset$ & $\emptyset$ & $\emptyset$ & $\emptyset$ & \textbf{1.00} & \textbf{1.00} & \textbf{1.00} & \textbf{1.00} & \textbf{1.00} \\ \hline

\multirow{3}{*}{\shortstack{EigenW.}} 
    & LR & \textbf{1.00} & 1.00 & \textbf{1.00} & $\emptyset$ & $\emptyset$ & $\emptyset$ & $\emptyset$ & $\emptyset$ & \textbf{1.00} & \textbf{1.00} & 0.88  & \textbf{1.00} &\textbf{ 1.00}  \\
    & SR$_a$ & 0.66 & \textbf{1.00} & 0.66 & $\emptyset$ & $\emptyset$ & $\emptyset$ & $\emptyset$ & $\emptyset$ & 0.67 & 0.61 & 1.00  & 0.68 & 0.60  \\\hline
	
\multirow{3}{*}{\shortstack{PenD.}} 
    & LR & \textbf{1.00} & \textbf{1.00 }& \textbf{1.00} & $\emptyset$ & $\emptyset$ & $\emptyset$ & $\emptyset$ & $\emptyset$ & \textbf{1.00 }& \textbf{1.00} & 0.55  & 0.94 & \textbf{1.00 } \\
    & SR$_a$ & 0.71 & 0.72 & 0.68 & $\emptyset$ & $\emptyset$ & $\emptyset$ & $\emptyset$ & $\emptyset$ & 0.67 & 0.70 & \textbf{0.99}  & \textbf{0.96} & 0.70  \\\hline
	
\multirow{3}{*}{\shortstack{AtrialF.}} 
    & LR & \textbf{0.86} & $\emptyset$ & $\emptyset$ & $\emptyset$ & $\emptyset$ & $\emptyset$ &  $\emptyset$ & \textbf{0.99} & $\emptyset$ & \textbf{0.92} & 1.00  & 0.97 & 0.58  \\
    & SR$_a$ & 0.67 & $\emptyset$ & $\emptyset$ & $\emptyset$ & $\emptyset$ & $\emptyset$ & $\emptyset$ & 0.67 & $\emptyset$ & 0.79 & 1.00  &\textbf{ 0.99} & \textbf{0.60 } \\\hline
	
\multirow{3}{*}{\shortstack{StandW.}} 
    & LR & \textbf{0.97} & $\emptyset$ & $\emptyset$ & $\emptyset$ & $\emptyset$ & $\emptyset$ & $\emptyset$ & 0.64 & $\emptyset$ & 0.50 & 1.00  &  0.67 & \textbf{0.99} \\ 
    & SR$_a$ & 0.67 & $\emptyset$ & $\emptyset$ & $\emptyset$ & $\emptyset$ & $\emptyset$ & $\emptyset$ & \textbf{0.70} & $\emptyset$ &  \textbf{0.61} & 1.00  &  \textbf{0.96} & \textbf{0.61} \\ \hline

\multirow{3}{*}{\shortstack{BasicM.}} 
    & LR & 0.54 & 0.54 & 0.54 & 0.55 & \textbf{0.54} & \textbf{0.54} & \textbf{0.54} & 0.54 & 0.57 & $\emptyset$ & 0.51 & 0.74 & 0.56 \\ 
    & SR$_a$ & \textbf{0.55} & \textbf{0.55} & \textbf{0.55} & \textbf{0.57} & 0.53 & 0.53 & 0.52 &\textbf{ 0.55} & \textbf{0.62} & $\emptyset$ & \textbf{0.99} & \textbf{0.97} & \textbf{0.74} \\ \hline

\multirow{3}{*}{\shortstack{Cricket}} 
    & LR & 0.81 & 0.75 & 0.74 & 0.71 & 0.74 & 1.00 & 0.78 & 0.77 & 0.72 & $\emptyset$ & 0.95 & 0.88 & 0.71 \\ 
    & SR$_a$ & \textbf{0.99} & \textbf{0.99} & \textbf{1.00} & \textbf{1.00} & \textbf{1.00} & \textbf{1.00} & \textbf{0.98} & \textbf{0.98} & \textbf{0.99} & $\emptyset$ & \textbf{1.00}  & \textbf{1.00} & \textbf{1.00} \\   \hline

\multirow{3}{*}{\shortstack{Epilepsy}} 
    & LR & 0.71 & \textbf{0.84} &\textbf{ 0.91} & 0.62 & 0.76 & \textbf{0.79 }& \textbf{0.89} & 0.65 & 0.53 & $\emptyset$ & 0.90 & 0.60 & 0.72 \\ 
    & SR$_a$ & \textbf{0.75} & 0.59 & 0.69 & \textbf{0.82 }& \textbf{0.80} & 0.70 & 0.63 & \textbf{0.71} & \textbf{0.74} & $\emptyset$ & \textbf{1.00} & \textbf{1.00} & \textbf{0.82} \\ \hline

\multirow{3}{*}{\shortstack{Handwriting}} 
    & LR & 0.85 & 0.61 & 0.65 & 0.57 & 0.57 & \textbf{0.92} & 0.63 & 0.57 & 0.57 & $\emptyset$ & 0.84 & 0.68 & 0.57 \\ 
    & SR$_a$ & \textbf{0.90} & \textbf{0.78} & \textbf{0.75} & \textbf{0.60} & \textbf{0.77} & 0.75 & \textbf{0.78} & \textbf{0.64} & \textbf{0.64} & $\emptyset$ & \textbf{1.00} & \textbf{1.00} & \textbf{0.74} \\ \hline
	
\multirow{3}{*}{\shortstack{Libras}} 
    & LR & 0.78 & 0.84 & 0.82 & 0.83 & 0.73 & 0.67 & 0.78 & 0.81 & \textbf{0.97} & $\emptyset$ & 0.71 & \textbf{0.80} & 0.95 \\ 
    & SR$_a$ & \textbf{0.89} & \textbf{0.93} & \textbf{0.98} & \textbf{0.90} & \textbf{1.00} & \textbf{0.87} & \textbf{0.93} & \textbf{0.95} & 0.84 & $\emptyset$ & \textbf{0.76} & 0.60 & \textbf{0.97} \\ \hline
	
\multirow{3}{*}{\shortstack{NATOPS}} 
    & LR & 0.93 & 1.00 & 0.97 & 1.00 & 1.00 & 0.93 & 0.99 & 1.00 & 1.00 & $\emptyset$ & 0.83 & 0.52 & 1.00 \\ 
    & SR$_a$ & \textbf{1.00} & 1.00 & \textbf{1.00} & 1.00 & 1.00 & \textbf{1.00} & \textbf{1.00} & 1.00 & 1.00 & $\emptyset$ & \textbf{1.00} & \textbf{1.00} & 1.00 \\ \hline
	
\multirow{3}{*}{\shortstack{RacketS.}} 
    & LR & 0.51 & 0.51 & 0.51 & 0.51 & 0.51 & 0.51 & 0.51 & 0.51 & 0.51 & $\emptyset$ & 0.98 & 0.93 & 0.51 \\ 
    & SR$_a$ & \textbf{0.61} & \textbf{0.66} & \textbf{0.65} & \textbf{0.65} & \textbf{0.69} & \textbf{0.64} & \textbf{0.66} & \textbf{0.67} & \textbf{0.67} & $\emptyset$ & \textbf{0.99} & \textbf{0.96} & \textbf{0.64} \\ \hline
	
\multirow{3}{*}{\shortstack{UWaveG.}} 
    & LR & 0.79 & 0.94 & 0.84 & 0.93 & 0.57 & 0.64 & 0.76 & 0.83 & 0.55 & $\emptyset$ & 0.86 & 0.72 & 0.57 \\ 
    & SR$_a$ & \textbf{0.97} & 0.94 & \textbf{0.96} & \textbf{0.96} & \textbf{0.97} & \textbf{0.89} & \textbf{0.93} & \textbf{0.95} & \textbf{0.97} & $\emptyset$ & \textbf{1.00} & \textbf{1.00} & \textbf{0.97} \\ \hline
	
\multirow{3}{*}{\shortstack{FingerM.}} 
    & LR & \textbf{0.94} & \textbf{1.00} & \textbf{1.00} &\textbf{ 1.00} & $\emptyset$ & $\emptyset$ & $\emptyset$ & \textbf{1.00} & \textbf{1.00} & \textbf{1.00} & $\emptyset$ & 0.85 & \textbf{1.00} \\ 
    & SR$_a$ & 0.74 & 0.92 & 0.90 & 0.90 & $\emptyset$ & $\emptyset$ & $\emptyset$ &  0.90 & 0.90 & 0.90 & $\emptyset$ & \textbf{0.96} & 0.90\\   \hline

\multirow{3}{*}{\shortstack{HandM.}} 
    & LR & 0.68 & 0.68 & 0.68 & 0.68 & $\emptyset$ & $\emptyset$ & $\emptyset$ & 0.68 & 0.68 & 0.68 & $\emptyset$ & 0.80 & 0.68 \\ 
    & SR$_a$ & \textbf{0.90} & \textbf{0.90} & \textbf{0.90} & \textbf{0.90} & $\emptyset$ & $\emptyset$ & $\emptyset$ &  \textbf{0.75} & \textbf{0.84} & \textbf{0.75} & $\emptyset$ & \textbf{0.97} &\textbf{ 0.91 }\\   \hline

\multirow{3}{*}{\shortstack{MotorI.}} 
    & LR & 0.66 & 0.66 & 0.56 & 0.57 & $\emptyset$ & $\emptyset$ & $\emptyset$ & \textbf{0.86} & \textbf{0.89} & \textbf{0.82} & $\emptyset$ & 0.98 & \textbf{0.83} \\ 
    & SR$_a$ & \textbf{0.70} & \textbf{0.71} & \textbf{0.60} & \textbf{0.60} & $\emptyset$ & $\emptyset$ & $\emptyset$ &  0.60 & 0.60 & 0.60 & $\emptyset$ & \textbf{1.00} & 0.60\\   \hline
	
\multirow{3}{*}{\shortstack{SelfR1.}} 
    & LR & 0.62 & 0.62 & 0.62 & 0.72 & $\emptyset$ & $\emptyset$ & $\emptyset$ & 0.62 & 0.60 & 0.59 & $\emptyset$ & 0.78 & 0.62 \\ 
    & SR$_a$ & \textbf{0.65} & \textbf{0.85} & \textbf{0.66} & \textbf{0.77} & $\emptyset$ & $\emptyset$ & $\emptyset$ &  0.62 & 0.60 & \textbf{0.61} & $\emptyset$ & \textbf{0.99} & \textbf{0.77}\\   \hline

\multirow{3}{*}{\shortstack{SelfR2.}} 
    & LR & \textbf{0.98} & 0.85 & 0.61 & \textbf{0.93} & $\emptyset$ & $\emptyset$ & $\emptyset$ & \textbf{0.92} & \textbf{0.96} & 0.88 & $\emptyset$ & 1.00 & \textbf{0.95} \\ 
    & SR$_a$ & 0.81 & \textbf{0.96} & \textbf{0.68} & 0.60 & $\emptyset$ & $\emptyset$ & $\emptyset$ &  0.83 & 0.83 & \textbf{0.91} & $\emptyset$ & 1.00 & 0.83 \\   \hline

\multirow{3}{*}{\shortstack{Heartbeat}} 
    & LR & 1.00 & 1.00 & $\emptyset$ & $\emptyset$ & $\emptyset$ & $\emptyset$ & $\emptyset$ & 0.93 & 0.94 & 0.94  & 0.75 & $\emptyset$ & 0.98 \\ 
    & SR$_a$ & 1.00 & 1.00 & $\emptyset$ & $\emptyset$ & $\emptyset$ & $\emptyset$ & $\emptyset$ & \textbf{0.96} & \textbf{0.97} & 0.94  &  \textbf{1.00} & $\emptyset$ & \textbf{ 1.00} \\  \hline

\multirow{3}{*}{\shortstack{ERing}} 
    & LR & 0.83 & 0.72 & 0.78 & $\emptyset$ & $\emptyset$ & $\emptyset$ & $\emptyset$ & 0.88 & 0.78 & 0.81 & 0.71 & 0.78  & $\emptyset$\\ 
    & SR$_a$ & \textbf{1.00} & \textbf{1.00} & \textbf{1.00} & $\emptyset$ & $\emptyset$ & $\emptyset$ & $\emptyset$ &\textbf{ 0.95} & \textbf{1.00} & \textbf{0.95} & \textbf{1.00} & \textbf{1.00} & $\emptyset$ \\   \hline

\multirow{3}{*}{\shortstack{EthanolC.}} 
    & LR & 0.99 & 0.99 & 0.99 & $\emptyset$ & $\emptyset$ & $\emptyset$ & $\emptyset$ & 0.99 & 0.99 & 0.99 & 0.99 & 0.99  & $\emptyset$\\ 
    & SR$_a$ & \textbf{1.00} & \textbf{1.00} & \textbf{1.00} & $\emptyset$ & $\emptyset$ & $\emptyset$ & $\emptyset$ & \textbf{1.00} & \textbf{1.00} & \textbf{1.00}  &\textbf{ 1.00} & \textbf{1.00} & $\emptyset$  \\ \hline  

\multirow{3}{*}{\shortstack{LSST}} 
    & LR & \textbf{0.79} & 0.58 & 0.58 & $\emptyset$ & $\emptyset$ & $\emptyset$ & $\emptyset$ & 0.58 & 0.57 & 0.57 & 0.51 & 0.58  & $\emptyset$ \\ 
    & SR$_a$ & 0.60 & 0.58 & \textbf{0.61} & $\emptyset$ & $\emptyset$ & $\emptyset$ & $\emptyset$ & \textbf{0.60} & \textbf{0.60} & \textbf{0.60} & \textbf{0.55} & \textbf{0.60}  & $\emptyset$ \\ \hline

\multirow{3}{*}{\shortstack{PEMS-SF}} 
    & LR & 0.77 & 0.94 & 0.84 & $\emptyset$ & $\emptyset$ & $\emptyset$ & $\emptyset$ & 0.92 & 1.00 & 0.95 & 0.97 & 0.99  & $\emptyset$\\ 
    & SR$_a$ & \textbf{1.00} & \textbf{1.00} & \textbf{0.97} & $\emptyset$ & $\emptyset$ & $\emptyset$ & $\emptyset$ & \textbf{1.00} & 1.00 & \textbf{1.00}  & \textbf{1.00} & \textbf{1.00} & $\emptyset$  \\ \hline  

\multirow{3}{*}{Wins(\%)} 
    & LR & \multicolumn{7}{c||}{22.3\%} & \multicolumn{6}{c|}{17.7\%}  \\ 
    & Ties & \multicolumn{7}{c||}{15.7\%} & \multicolumn{6}{c|}{20.0\%}  \\ 
    & SR$_a$ & \multicolumn{7}{c||}{62.0\%} & \multicolumn{6}{c|}{62.3\%}  \\  
    \hline
\end{tabular}
}
\end{table*}
\clearpage
\noindent \textbf{Inference runtime comparison of the different OOD detection algorithms.}
We provide in Tables \ref{tab:params} and \ref{tab:runtime} a comparison of the number of parameters and the runtime between different OOD detection methods for time-series. Intuitively, both HL and SR methods are characterized by a larger number of parameters than LL and LR as the latter two methods only rely on a single VAE model to compute the OOD score. However, we can observe that LR has the longest score computation runtime: this is due to the new training iterations LR introduces to compute the OOD score of each example. On the other hand, SR algorithm only runs a single inference pass for each example, then computes the ratio between both computed likelihoods. This approach of SR algorithm yields a fast and accurate OOD detector.
\begin{table*}[!h]
\centering
\setlength\extrarowheight{2pt}
\caption{Number of parameters of each DNN used by the different OOD methods.}
\label{tab:params}
\begin{tabular}{|c|c|}
\cline{2-2}
  \multicolumn{1}{c|}{ } & \textbf{Number of parameters}   \\ \hline
     LL & 454,628  \\ \hline
     HL & 687,268 \\ \hline
     LR & 454,628 \\ \hline
     SR & 909,256 \\ \hline
\end{tabular}
\end{table*}

\begin{table*}[!h]
\centering
\setlength\extrarowheight{2pt}
\caption{OOD Inference runtime comparison on the different datasets using different OOD methods.}
\label{tab:runtime}
\begin{tabular}{|c|cccccc|}
\cline{2-7}
  \multicolumn{1}{c|}{ } & \multicolumn{6}{c|}{\textbf{Runtime (seconds)}}   \\ \cline{2-7}
 \multicolumn{1}{c|}{ } & AWR & SWJ & CkT & HMD & Hbt & ERg   \\ 
 \multicolumn{1}{c|}{ } & (Motion) & (ECG) & (HAR) & (EEG) & (Audio) & (Other)   \\ \hline  
 LL & 0.66 & 0.32 & 0.43 & 0.49 & 0.49 & 0.22  \\ \hline
 HL & 1.06 & 0.65 & 0.55 & 0.81 & 0.90 & 0.47  \\ \hline
 LR & 3.72 & 1.48 & 1.40 & 1.35 & 1.38 & 1.42  \\ \hline
 SR & 1.5 & 0.66 & 0.80 & 1.01 & 1.02 & 0.52  \\ \hline
\end{tabular}
\end{table*}

\clearpage
\section{Summary and Future Work}

We introduced a novel seasonal ratio (SR) score to detect out-of-distribution (OOD) examples in the time-series domain. SR scoring relies on Seasonal and Trend decomposition using Loess (STL) to extract class-wise semantic patterns and remainders from time-series signals; and estimating class-wise conditional likelihoods for both input time-series and remainders using deep generative models. The SR score of a given time-series signal and the estimated threshold interval from the in-distribution data enables OOD detection. Our strong experimental results demonstrate the effectiveness of SR scoring and alignment method to detect time-series OOD examples over prior methods. Immediate future work includes applying seasonal ratio score based OOD detection to generating synthetic time-series data for small-data settings.

\section{Acknowledgments}

The authors would like to thank Alan Fern for the useful discussions regarding the key assumption behind the seasonal ratio scoring approach. This research is supported in part by the AgAID AI Institute for Agriculture Decision Support, supported by the National Science Foundation and United States Department of Agriculture - National Institute of Food and Agriculture award \#2021-67021-35344.

\clearpage
\bibliographystyle{apalike}
\bibliography{sample-base}

\end{document}